\documentclass[review]{elsarticle}
\graphicspath{ {./figures/} }
\usepackage{hyperref}
\usepackage{float}
\usepackage{verbatim} 
\usepackage{apalike}
\restylefloat{figure}
\floatstyle{plaintop} 
\restylefloat{table}

\usepackage{hhline} 
\usepackage{enumitem}

\usepackage{xcolor}

\usepackage{graphicx}
\usepackage{lscape}

\journal{Expert Systems with Applications}

\biboptions{authoryear}

\begin{document}
\begin{frontmatter}


\begin{titlepage}
\begin{center}
\vspace*{1cm}

\textbf{ \large Overtake Detection in Trucks Using CAN Bus Signals: A Comparative Study of Machine Learning Methods}

\vspace{1.5cm}

Fernando Alonso-Fernandez (feralo@hh.se), Talha Hanif Butt (talha-hanif.butt@hh.se), Prayag Tiwari (prayag.tiwari@hh.se) \\

\hspace{10pt}

\begin{flushleft}
\small  
School of Information Science, Computer and Electrical Engineering, Halmstad University, Box 823, Halmstad SE 301-18, Sweden 

\vspace{1cm}
\textbf{Corresponding author:} \\ 
Fernando Alonso-Fernandez \\
School of Information Science, Computer and Electrical Engineering, Halmstad University, Box 823, Halmstad SE 301-18, Sweden  \\
Tel: +46 729 77 35 29 \\
Email: feralo@hh.se

\end{flushleft}        
\end{center}
\end{titlepage}

\title{Overtake Detection in Trucks Using CAN Bus Signals: A Comparative Study of Machine Learning Methods
}


\author{Fernando Alonso-Fernandez\corref{cor1}}
\ead{feralo@hh.se}

\author{Talha Hanif Butt}
\ead{talha-hanif.butt@hh.se}

\author{Prayag Tiwari}
\ead{prayag.tiwari@hh.se}

\cortext[cor1]{Corresponding author.}
\address{School of Information Science, Computer and Electrical Engineering, Halmstad University, Box 823, Halmstad SE 301-18, Sweden }

\begin{abstract}

Safe overtaking manoeuvres in trucks are vital for preventing accidents 
and ensuring 
efficient traffic flow. 
Accurate prediction of such manoeuvres is essential for Advanced Driver Assistance Systems (ADAS) to make timely and informed decisions. 
In this study, we focus on overtake detection using Controller Area Network (CAN) bus data collected from five in-service trucks provided by the Volvo Group. 
%
%
%
We evaluate three common classifiers for vehicle manoeuvre detection, Artificial Neural Networks (ANN), Random Forest (RF), and Support Vector Machines (SVM), and analyse how different preprocessing configurations affect performance.
%
%
%
We find that variability in traffic conditions strongly influences the signal patterns, particularly in the no-overtake class, affecting classification performance if training data lacks adequate diversity. 
%
%
Since the data were collected under unconstrained, real-world conditions, class diversity cannot be guaranteed a priori. However, training with data from multiple vehicles improves generalisation and reduces condition-specific bias.
%
%
Our per-truck analysis also reveals that classification accuracy, especially for overtakes, depends on the amount of training data per vehicle. 
%
%
To address this, we apply a score-level fusion strategy, which yields the best per-truck performance across most cases.
%
%
Overall, we achieve an accuracy via fusion of \textit{TNR}=93\% (True Negative Rate) and \textit{TPR}=86.5\% (True Positive Rate). 
This research has been part of the BIG FUN project, which explores how Artificial Intelligence can be applied to logged vehicle data to understand and predict driver behaviour, particularly in relation to Camera Monitor Systems (CMS), being introduced as digital replacements for traditional exterior mirrors.
%
%
%
%
%
%
%
%
%
\end{abstract}


\end{frontmatter}

\section{Introduction}

Developing Advanced Driver Assistance Systems (ADAS) is a major focus in artificial intelligence research for the automotive industry. 
These systems use data from various sensors, such as cameras, LiDAR, radar, biosensors, and vehicle networks, to alert drivers of potential hazards or autonomously adjust vehicle behaviour to prevent accidents \citep{badue2021self, kumar2020recent}. 
A key requirement for effective ADAS is the ability to analyse environmental and vehicular data, including speed, acceleration, and lane position, for both the ego and surrounding vehicles \citep{hasan2021multi}. 
Additionally, many systems must assess the driver's intention in order to issue timely warnings or override inputs if needed \citep{xing2019driver, zhang2021intention}.
Understanding driver behaviour in contexts such as overtaking, braking, turning, or lane changes is thus essential to improving ADAS performance and vehicle safety \citep{mozaffari2020deep}. 
Such insights support systems responsible for lane departure warning, blind-spot detection, collision avoidance, and adaptive cruise control, among others.

In this work, we focus on the design and evaluation of an automatic overtake detection system for trucks. 
Compared to other vehicles, trucks pose greater risks due to their size and mass, making truck-related accidents potentially more severe. 
These accidents can lead to more fatalities, traffic congestion, and significant economic loss due to cargo delays \citep{fu2021truck, mcknight2020factors}.
%
%
%
%
To address this and contribute to improved safety and efficiency in freight transportation, we aim to enhance the ability to detect and monitor overtaking behaviour in trucks using CAN bus signals. 
%
%
%
These signals are available onboard without requiring additional hardware, such as cameras or biosensors \citep{garcia2020can, neumeier2019impact}, making them a cost-effective and privacy-friendly option for deployment. 
This avoids concerns related to video-based monitoring and supports scalable ADAS solutions for heavy-duty vehicles.
%
%
%

For this study, we utilise CAN data collected from actual, in-service trucks, generously provided by the Volvo Group, a key research partner in this work. 
This research has been conducted within the framework of the BIG FUN project (Big Data-Powered End User Function Development), which explores how Artificial Intelligence (AI) can be applied to logged vehicle data to identify and predict driver behaviour, particularly in relation to Camera Monitor Systems (CMS), which are being introduced as replacements for traditional exterior mirrors. While CMS technologies offer clear advantages in terms of visibility and aerodynamics, they may also influence driver behaviour in ways that are not yet fully understood. In addition, they provide new opportunities for analysing logged data, as these systems inherently generate digital video, which can be combined with other sensor data streams. Overtaking manoeuvres are especially relevant in this context, as indirect vision, whether provided by CMS or traditional mirrors, plays a critical role in driver decision-making during such actions. Motivated by this, our study focuses on the detection of overtaking events using CAN bus data, which is available both in new CMS trucks and in older vehicles with conventional mirrors. Our long-term aim is to enable behavioural analysis and comparison between the two configurations, as drivers adapt to CMS and begin to rely on digital video instead of traditional optical systems.
%

Our overtake detection approach using CAN data, built on popular machine learning techniques, extends methods from conclusions and recommendations from related literature on vehicle manoeuvre detection \citep{xing2019driver}. 
Although we use standard classifiers like ANN, RF, and SVM, our primary contribution lies in applying them in what is, to the best of our knowledge, one of the first studies on overtake detection in trucks using real CAN bus data. While most previous works rely on a single classifier, we systematically compare these well-established, yet diverse models under the same experimental conditions and analyse how their performance is influenced by different preprocessing configurations. This also allows to explore the benefits of classifier fusion, which, as our results show, contributes to improved robustness across different trucks.
%
%
Our work also encompasses collecting and annotating a dataset to enable training and evaluation of the system.
Most prior works deal with other manoeuvres (e.g., lane changes) or use simulated or passenger vehicle data, which may not generalise to the context of commercial trucks. 
%
%
Here, we present an original dataset of annotated real-world CAN signals from five operational trucks in naturalistic driving scenarios.
As illustrated in Figure~\ref{fig:system}, several CAN signals, which measure various aspects of vehicle operation (such as pedal positions, speed, etc.), are continuously logged. 
When a predefined trigger condition is met (indicated by an arrow in the figure), a crop of these signals is extracted and passed to a classifier, which determines whether the crop corresponds to an overtake. 
The signals in each crop are usually pre-processed using a sliding window approach, where metrics such as the average and standard deviation of the signals are computed within the window. 

A preliminary version of this work was presented at a conference \citep{Butt2024}, where a smaller dataset (712 files from 3 trucks) was employed. 
In this paper, we expand the dataset to 1247 files, increasing the number of trucks analysed from 3 to 5. 
Additionally, the previous work employed a fixed configuration for the crop and sliding window sizes. 
Here, we conduct an in-depth investigation on how different preprocessing and crop configurations affect detection.
In particular, we analyse the influence of the crop size, sliding window size, and metrics computed within each window. 
This analysis is motivated by the absence of studies dedicated to overtake detection in trucks and the lack of consensus on optimal configurations of these parameters in related research, as discussed in 
\citep{xing2019driver,khairdoost2020real},
and observed by the different parameter choices in works such as 
\citep{kim2017prediction,zhang2020hybrid,khairdoost2020real}.
Some studies even consider the use of dynamic crops to cope with manoeuvres with varying time lengths
\citep{zheng2017lane,das2020detecting,das2023lanechange}.
The present paper also provides insights missing in the previous version, such as the impact of the number of trucks used to train the classifiers, the visual patterns of the different CAN signals, or the per-truck performance of the system. 
Finally, we demonstrate that classifier fusion can enhance per-truck accuracy, improving system generalisation and robustness.

%

The rest of the paper is organised as follows. Section~\ref{sect:soa} describes related work. Section~\ref{sect:material_methods} extensively describes the proposed system, including the database, machine learning methods, and experimental protocol. The experimental results are presented and discussed in Section~\ref{sect:results}. Finally, the conclusions are given in Section~\ref{sect:conclusions}.

\begin{landscape}
\begin{table}[]
\centering
\caption{Summary of related works on overtake, lane change and turning detection using CAN signals. OV=Overtake. LC=Lane Change. LK=Lane Keep. TU=Turning. GS=Going Straight. Other acronyms are defined in the text. }
\label{tab:summary-related-works}
\resizebox{\columnwidth}{!}{%
\begin{tabular}{llllllll}

\multicolumn{8}{}{} \\

\textbf{Task} & \textbf{Work} & \textbf{Vehicle} & \textbf{\begin{tabular}[c]{@{}l@{}}CAN\\ signals\end{tabular}} & \textbf{\begin{tabular}[c]{@{}l@{}}Preprocessing of\\ CAN signals\end{tabular}} & \textbf{Classifier} & \textbf{Data} & \textbf{\begin{tabular}[c]{@{}l@{}}Results\\ (CAN only)\end{tabular}} \\ \hline \hline

 & \citep{blaschke2008predicting} & Car & 5 & Raw signals & Fuzzy Logic  & 43 OV, 55 no OV & TPR=95\%, TNR=93\%  \\

 &  &  &  &  &  & 156 OV, 16 no OV & TPR=94.2\%, TNR=100\% \\ \cline{2-8}
 
OV & \citep{stefansson2020modeling} & Simulator & 2 & Raw signals & Kalman filter & 52 OV & n/a  \\ \cline{2-8}
 
 & \citep{lin2021implementation} & Golf cart & 5 & Raw signals & Vision & n/a & n/a  \\ \cline{2-8}
 
 & \textbf{This work} &  \textbf{Truck} & 10 & \begin{tabular}[c]{@{}l@{}}Rule for OV candidates, \\  then mean, variance. \\ \textbf{Several crop/window sizes}\end{tabular} & ANN, RF, SVM  & 382 OV, 865 no OV & \begin{tabular}[c]{@{}l@{}}TPR=86.5\%, TNR=93\% (SVM+RF)\\ TPR=86.2\%, TNR=95.1\% (RF)\end{tabular}  \\ \hline \hline

 &  \citep{kim2017prediction} & Simulator & 11 & \begin{tabular}[c]{@{}l@{}}Mean, variance, PCA. \\ Crops=0.5 sec, window=0.2 sec\end{tabular} & SVM  & n/a & Accuracy=90-97\%  \\  \cline{2-8}

 &  \citep{zheng2017lane} &  Car & 4 & \begin{tabular}[c]{@{}l@{}}LC candidates and TU removal \\ via steering angle information \end{tabular} & \begin{tabular}[c]{@{}l@{}}DTW+KNN\\ HMM\end{tabular} &  \begin{tabular}[c]{@{}l@{}}UTDrive dataset\\ 122/209 left/right LC, 87/404 left/right TU, 2943 LK \end{tabular} & Accuracy=58-92\%  \\  \cline{2-8}

LC &  \citep{das2020detecting} &  Car & 4 & \begin{tabular}[c]{@{}l@{}}LC candidates using cameras, \\ then mean, max, min, std \end{tabular} & \begin{tabular}[c]{@{}l@{}}RF, SVM\\ ANN, XGBoost\end{tabular}  & \begin{tabular}[c]{@{}l@{}}SHRP2 dataset\\ 1200 LC, 2400 no LC\end{tabular} & \begin{tabular}[c]{@{}l@{}}AUC=0.81-0.864\\ Accuracy=74.7-76.3\%\end{tabular}  \\  \cline{2-8}
 
 &  \citep{guo2022lane} &   Car & 2 & \begin{tabular}[c]{@{}l@{}}Sliding window prediction\\ every 0.1 sec\end{tabular} & \begin{tabular}[c]{@{}l@{}}Autoencoder,\\ Transformer\end{tabular} & \begin{tabular}[c]{@{}l@{}}DAS1 dataset\\ 2005/1480 left/right LC, 34850 LK\end{tabular} & \begin{tabular}[c]{@{}l@{}}n/a (combined vision, \\ GPS and CAN)\end{tabular}  \\ \cline{2-8}
 
 &  \citep{das2023lanechange} &  Car & 4 & \begin{tabular}[c]{@{}l@{}}Same as \citep{das2020detecting}\\ Class balance (SMOTE, RMUS)\\ XGboost for correlation removal\end{tabular} & \begin{tabular}[c]{@{}l@{}}ResNet18 CNN\\ ANN, XGBoost, CART\\ SVM, RF, KNN, NB\end{tabular}& \begin{tabular}[c]{@{}l@{}}SHRP2 dataset\\ 1200 LC, 68173 no LC\end{tabular} & \begin{tabular}[c]{@{}l@{}}CNN: Recall=82\%, Accuracy=77.9\%\end{tabular}  \\  \hline \hline
 
 & \citep{Huang2019turning} &   Car & 2 & Raw signals & FCN+GMM  & n/a & \begin{tabular}[c]{@{}l@{}}n/a (combined vision, \\ GPS and CAN)\end{tabular}  \\ \cline{2-8}

TU & \citep{zhang2020hybrid} &   Car & 4 & \begin{tabular}[c]{@{}l@{}}Signals estimated from video\\ Crops 11 sec, window=0.1 sec\end{tabular} & Bi-LSTM  & \begin{tabular}[c]{@{}l@{}}NGSIM dataset\\ 1039/978 left/right TU, 976 GS\end{tabular} & \begin{tabular}[c]{@{}l@{}}Accuracy=74.5, 93.5, 94.2\%\\ (3, 2, 1 sec ahead TU)\end{tabular}  \\ \cline{2-8}
 
 & \citep{khairdoost2020real} &  Car & 10 & \begin{tabular}[c]{@{}l@{}}Histogram, min, avg, max\\ Crop=2/3 sec\end{tabular} & LSTM & \begin{tabular}[c]{@{}l@{}}RoadLAB dataset\\ 65/40 left/right LC, 65/75 left/right TU, 80 GS\end{tabular} & \begin{tabular}[c]{@{}l@{}}n/a (combined vision, \\ driver gaze/head and CAN)\end{tabular}  \\ \hline

\end{tabular}%
}
\end{table}
\end{landscape}

\section{Related Work}
\label{sect:soa}




%
%
The overtaking manoeuvre is one of the most critical and challenging driving actions, involving a sequence of decisions such as lane changes, acceleration and deceleration, and real-time estimation of the speed and distance of surrounding vehicles. 
Despite its importance, research focusing on overtaking prediction using Controller Area Network (CAN) data, especially in trucks, is scarce. 
The few existing studies are focused on other vehicle types, such as passenger cars or experimental prototypes \citep{blaschke2008predicting, stefansson2020modeling, lin2021implementation}. 
Because overtaking carries a high risk, no real-world dataset is available either \citep{dutra2022data}. 
The majority of related research concentrates on detecting other types of manoeuvres, such as turning intention at intersections 
or lane change prediction, 
a manoeuvre closely related to overtaking.
Detecting these manoeuvres is done using diverse data sources \citep{xing2019driver} which measure: 

\begin{enumerate}[label=\roman*)]
    \item The vehicle dynamics, engine and driver inputs, such as speed, acceleration, pedals' position, wheel angle, etc. This is commonly captured by the Electronic Control Units (ECUs) of the vehicle. 
    
    \item The traffic information, such as position, distance, and motion of surrounding vehicles, or other road users like cyclists and pedestrians, as well as traffic signs and road hazards. This is typically measured by sensors such as radar, LIDAR, cameras, or ultrasonic sensors. Researchers may also employ GPS or map information, or include aspects such as temperature, weather or road condition \citep{kim2017prediction}.
    
    \item The vehicle in relation to the surrounding environment, such as the status and warnings regarding lane departure, blind spots, collisions, etc. These measurements are usually enabled by the data sources of $i$) and $ii$). 
    
    \item The driver, such as gaze direction, head position, fatigue levels, drowsiness, distraction, etc. This is captured via eye trackers, cameras or biosensors that capture physiological data such as EEG or ECG. In some cases, studies include demographic information of the driver, such as age, gender, driving experience, etc., although this information may proceed as part of annotated data collection, rather than actual automatic estimation \citep{das2020detecting}. 
    
\end{enumerate}

The Controller Area Network (CAN) is a communication protocol that aggregates vehicle data to enable its efficient real-time operation and coordination of its subsystems (engine, transmission, braking, etc.). It is also used for vehicle diagnostics, maintenance, and performance monitoring.
It typically captures aspects related to $i$-$iii$) above, and even if these can engage sensors such as cameras, radar, or LIDAR, the data transmitted in the CAN protocol only contains processed outputs from these sources (e.g., distances to obstacles or their velocity) rather than the raw sensor streams. 
We contrast this with other research which explicitly uses such raw data streams for manoeuvre detection, e.g. \citep{xing2019driver}.
Since they are not readily available from CAN data, external sensors placed inside or outside the vehicle are usually engaged.
In this section, we thus review relevant studies emphasising the utilisation of CAN data only.
Table~\ref{tab:summary-related-works} presents a summary of the relevant works covered in this Section.

Another related field of interest is the intention recognition of surrounding vehicles in intelligent and autonomous systems. These approaches aim to assess, for example, whether it is an appropriate time for the ego vehicle to perform a manoeuvre
such as changing lanes 
\citep{wang2022lanechangeautonomous}
or overtaking \citep{ORTEGA2024100264} based on surrounding traffic conditions.
Similarly, they attempt to predict the intentions of nearby vehicles, such as whether another vehicle will change lanes 
\citep{Liu2023lanechangesurrounding}, 
cross or turn at an intersection \citep{Pourjafari2024intersections},
or initiate an overtaking manoeuvre \citep{Lin2020overtake}.
%
%
However, we view this line of research as following a different focus.
Rather than detecting manoeuvres being executed by the current vehicle, these systems infer the future actions of other road users. 
In addition, CAN data primarily captures internal vehicle information, so it could provide very little evidence of such phenomena.
As a result, such approaches typically rely on external perception data from cameras, LiDAR, or radar to observe the environment. 
%
%
%
Therefore, we consider those approaches on intention recognition of other vehicles to be outside the scope of this paper.

\subsection{Overtake detection}

As mentioned earlier, the literature on overtaking detection is limited. Most studies discussed in this section focus on assessing whether overtaking is safe. We include them primarily because they utilise CAN data to analyse the manoeuvre.


\citep{blaschke2008predicting} presented one of the first studies focused on predicting overtaking manoeuvres using CAN bus data. 
In a first stage, they involved 28 drivers between 22-65 years old on country roads.
A vehicle ahead drove partly at the same speed and partly 30 km/h slower than the recommended speed limit. 
The data generated included 43 overtaking and 55 no overtaking manoeuvres, achieving an accuracy in detecting these two manoeuvres of TPR=95\% and TNR=93\%, respectively. 
Their fuzzy logic approach integrated 5 indicators to predict overtaking behaviour (distance and speed difference to the preceding vehicle, brake pressure, accelerator pedal pressure and accelerator pedal speed). In addition, the algorithm is applied only to time points with a speed above 60 km/h.
In a second stage, they obtained 156 overtake and 16 no overtake manoeuvres on country roads, motorways and city traffic, achieving TPR=94.2\% and TNR=100\%.
An observed limitation compared to our present study is the 
relatively small dataset.

\citep{stefansson2020modeling} proposed models for the decision-making process of human drivers.
The authors introduced a mathematical formulation of the overtaking problem and then proposed decision models judging whether a vehicle should overtake or not based on inputs such as a slow-moving vehicle ahead or an approaching vehicle in the other direction on a two-way road. 
The decision models were based on a Kalman filter estimator with distance and velocity difference between vehicles as input signals, and two decision rules were derived from different risk-awareness levels.
An experimental testbed was also introduced using a driving simulator with two drivers carrying out 52 overtaking decisions, classifying the manoeuvre as safe or unsafe.

\citep{lin2021implementation} proposed a Model Predictive Control (MPC) scheme integrated with time-to-lane-crossing (TLC) estimation for autonomous vehicles (an electric prototype golf cart). 
Thus, the system is not really a detector of the driver's intention, but a detector of when an autonomous vehicle can safely overtake others.
They used 5 CAN bus signals for communication between sensors, actuators, and the vehicle control unit (VCU), but the system relied heavily on a vision-based system to detect lane lines and other vehicles to estimate TLC and decide whether to overtake or stay in lane. This limits its applicability to pure CAN data environments. 
The system determined the optimal throttle, brake, and steering commands to ensure safe longitudinal and lateral vehicle control. 

In the present paper, we used 10 onboard CAN signals to predict overtakes with three machine learning methods, ANN, RF and SVM. 
While these classifiers are popular choices in vehicle manoeuvre detection, appearing in several studies of Table~\ref{tab:summary-related-works}, our study is, to the best of our knowledge, the first to focus specifically on trucks using real-world CAN data from operational heavy-duty vehicles. 
Unlike most previous studies that rely on a single classifier, we compare multiple models and also evaluate their fusion.
Data were collected from five trucks, comprising 382 overtake and 865 no-overtake segments.
A precondition rule was used to identify potential overtakes, which were then manually annotated via video review to enable classifier training and evaluation. After such data labelling, video data was not further considered, so the developed predictors could rely only on CAN data.
Unlike previous works that used fixed crops and sliding window sizes, and given the lack of consensus regarding their optimal values \citep{xing2019driver,khairdoost2020real}, here we systematically explored multiple configurations to assess their impact on performance, and found that the optimal configuration varies depending on the classifier used.
Our best results overall were obtained with RF (TPR=86.2\%, TNR=95.1\%), while fusion with SVM improved robustness across vehicles, achieving TPR=86.5\% and TNR=93\%.

\subsection{Lane change detection}


The majority of related studies are in the context of lane change prediction.
The task is usually modelled using left lane change, right lane change, and lane keeping as classification classes. 
Detecting such manoeuvres can prevent accidents in scenarios with limited visibility, heavy traffic, or when merging, enabling timely alerts to avoid potential collisions.
%
%
\citep{xing2019driver} carried out a state-of-art literature review, focusing on highways.
%
%
%
Being from 2019, the studies reported mostly correspond to the pre-deep-learning era, with algorithms covering a wide range of methods, including Support Vector Machines (SVM), Hidden Markov Models (HMM), Artificial Neural Networks (ANN) and others, and just one study applying Long Short-Term Memory (LSTM) networks.
The obtained accuracies vary, with some studies surpassing 90\% and reaching up to 98-100\%. The prediction horizon was, at maximum, 3.5 seconds before the lane change, with many studies employing 1 second or less. 

Among the studies reported in \citep{xing2019driver}, very few used only vehicle information from CAN signals as data input, with many relying on other dedicated sensors to capture the data aspects $i$-$iv$) mentioned at the beginning of this section. Some do not even use CAN data at all.
%
%
Regarding databases, some works employed naturalist on-road data, while others employed a driving simulator. However, the data is unavailable in general, making replication difficult. 
In what follows, we report more recent studies on lane change prediction that are not included in the mentioned literature review paper.

\citep{kim2017prediction} used different combinations of 11 onboard signals to detect a lane change with an SVM classifier and data captured with a PC driving simulator. 
The signals were processed with a time crop and an overlapping sliding window of 0.5 and 0.2 seconds, respectively. 
As features, they used the average, variance and principal components (PCA) of the windowed signals.
The simulation contained different road conditions (dry, gravel, wet and snowy), also estimated from the signals with an ANN.
Accuracies in lane change detection varied between 90 and 97\% with the best-observed combination of signals.

\citep{zheng2017lane} applied two machine learning methods, one based on Dynamic Time Warping (DTW) features with a k-Nearest Neighbor (k-NN) classifier and another using an HMM classifier.
They considered four CAN signals, including vehicle speed, steering angle, brake pressure, and engine rpm.
Data came from a subset of the UTDrive naturalistic driving corpus, comprising 14 females and 44 males driving a Toyota RAV4 vehicle, with a mix of local and residential roads having 2-3 lanes and an average speed of 40 mph.
In the first stage, left/right turn manoeuvres are filtered out by seeking very large steering angle changes.  
Then, potential boundaries between lane-keeping (straight wheels) and lane-changing (non-straight wheels) are sought by applying a filter bank to the spectrogram of the steering angle signal. This is because a high variation of energy across bands has been observed to be indicative of the beginning of a lane change.
Afterwards, candidate lane change segments are classified by the k-NN and HMM methods.
The driving sessions of the database accounted for a wide variability of speeds, so a lane change manoeuvre may be executed over varying time lengths. The proposed spectral segmentation allowed accounting for such variability, which makes the use of fixed window segments undesirable.
In addition, DTW and HMM are also inherently capable of coping with data sequences of different lengths.
The obtained accuracies ranged between 58 and 92\%, depending on the class and the algorithm.

\citep{das2020detecting} also employed non-uniform time crops of the signals. 
To do so, they segmented 1200 lane-change segments from the SHRP2 Naturalistic Driving Study (NDS) dataset with a vision technique that analyses the offset of the vehicle with respect to the lane centre since the database not only contains CAN data. 2400 no lane-change segments were also extracted as negative class.
The database contains a variety of weather conditions (clear, snow, rain, and fog).
Lane change classification with four CAN signals was done by comparing RF, SVM, ANN, and Extreme Gradient Boosting (XGBoost) algorithms.
The dataset included four vehicle signals: speed, longitudinal acceleration, lateral acceleration, and yaw rate.
To account for variability in the manoeuvre duration, the mean, maximum,
minimum, and standard deviation of the signals in the segments were used as features.
With only CAN signals, the results obtained were AUC=0.864 (RF), 0.855 (SVM), 0.81 (ANN), and 0.859 (XGBoost) 
%

Deep learning methods have increasingly been applied in recent years for lane change detection, although the literature employing CAN signals is very scarce \citep{guo2022lane,das2023lanechange}, being more popular the use of video data \citep{scheel2019lanechange,XIE201941,nalcakan2023lanechange} including drones \citep{xu2021lanechangedrones,chen2021lanechangedrones}, or video combined with GPS information \citep{xing2020lanechange}.
%

\citep{guo2022lane} applied an autoencoder-based lane change detector and a transformer-based lane change predictor.
The employed DAS1 database, which is public, included 7960 trips recorded by 98 sedans, with data coming from a forward-looking camera integrated into an after-market ADAS system (Mobileye), a GPS unit, and CAN signals.
However, among the 17 signals available, only two are CAN signals (yaw rate and longitudinal acceleration), and the accuracy with only those or their impact was not explicitly assessed.
%
%

The authors of \citep{das2020detecting} extended the study using a ResNet-18 Convolutional Neural Network (CNN) \citep{das2023lanechange}. They first balanced the classes using synthetic minority oversampling (SMOTE) and random majority undersampling (RMUS), and then applied XGBoost to remove highly correlated features. 
Then, they converted the time sequence signals to an image, which was given to the CNN.
With only CAN signals, the reported recall was 82\% and the overall accuracy of the two classes was 77.9\%.
The CNN method was also observed to outperform other ML models in terms of recall, although precision was lower in some cases.

\subsection{Turning detection}


Turn intention prediction is another field of research that is related.
The task is usually addressed using left turn, right turn and going straight as classes. 
Predicting such intentions is crucial in intersections, ramps, parking areas, and roundabouts, where reduced visibility and potential collisions with other road users pose significant risks.

\citep{Huang2019turning} used a combination of multiple signals from CAN data, front and side cameras, GPS and an Inertial Measurement Unit (IMU) placed on a  Toyota Lexus sedan driven for 30 hours during 14 trips in different weather and lighting conditions. 
The CAN signals considered included steering wheel angle and gas pedal value, which allows the acceleration or turning rate to be modelled.
Different expert predictors were integrated, including a VGG CNN for image data and fully connected networks (FCN) for the remaining data.
The output of the experts is concatenated and sent to another fully connected network, which outputs Gaussian Mixture Models (GMM) parameters.
The paper studied the complementarity of the expert predictors, and although it did not evaluate the use of CAN data separately, dropping CAN signals was seen to produce an increase in the prediction error of 13.29\%.

\citep{zhang2020hybrid} developed a hybrid model that first applied autoregressive integrated moving average (ARIMA) to estimate the lateral position, longitudinal position, speed, and acceleration of the vehicle from video data, followed by a bidirectional LSTM (Bi-LSTM) network to predict turning behaviour using the estimated signals.
They used the open-source NGSIM dataset, which contains both video data and the mentioned vehicle signals as a reference, although the latter was estimated by the authors of the database from the video stream as well. 
They employed a total of 2993 samples of data containing left/right turns and going straight segments, obtained by analysing the heading angle of the vehicle, and then extracted a crop of 11 seconds of the entire turning process. 
The method is evaluated with a sliding window approach having a length of 1.5 seconds and a moving step of 0.1 seconds.
The obtained average recognition rate was 74.5\%, 93.5\% and 94.2\% at 3, 2 and 1 seconds, respectively, before initiating the manoeuvre.

\citep{khairdoost2020real} applied LSTM to a combination of driver gaze, head position, a forward vision system on the rooftop, and CAN data from 325 segments of the RoadLAB 
database, captured using an equipped sedan driven by 16 drivers on a 28.5 km pre-determined city course that included urban and suburban areas.
The study considered feature aggregation across 20 frames of data captured at 30 Hz (2/3 of a second).
%
%
For each crop, CAN data consisted of the histogram of steering wheel angles, the binary value of left and right turn signals, and the minimum, average and maximum vehicle speed, brake
pedal pressure, gas pedal pressure and speeds of the four
wheels. 
%
%
The precision/recall was 85.6\% and 84.1\%, but with the entire set of signals, not providing results with CAN data only.
The study also considered the use of left/right lane change as possible additional manoeuvres, with a precision/recall over the extended set of classes of 84.2\% and 82.9\%.

\begin{figure} [htb]
\centering
\includegraphics[width=0.95\textwidth]{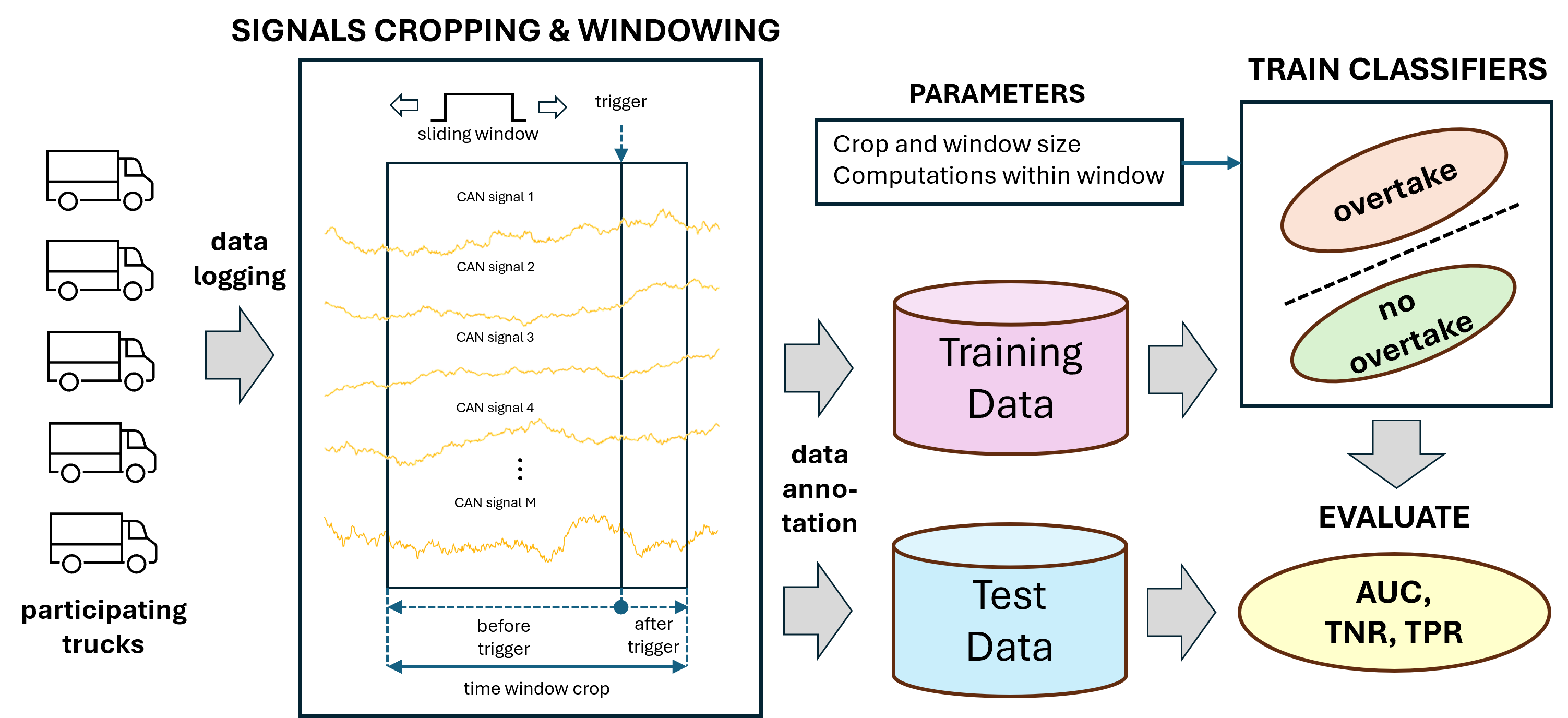}
\caption{Overview of the overtake classification system.}
\label{fig:system}
\end{figure}

\section{Materials and Methods}
\label{sect:material_methods}

This section describes the composition of our dataset, the machine learning methods and the experimental protocol employed.
An overview of our overtake classification system is shown in Figure~\ref{fig:system}.
We follow standard practices commonly used in manoeuvre detection studies, including data selection based on trigger conditions, manual annotation, signal preprocessing, model training, and performance evaluation using standard metrics such as Area Under the Curve (AUC), TNR and TPR. 
The manual annotation of overtake events was conducted through synchronised video review to ensure reliable ground truth labels.
%
%
To validate our approach, we use a standard hold-out evaluation, where classifiers are trained on a predefined training set until convergence and then tested on a separate, disjoint test set, ensuring that the evaluation reflects generalisation to unseen data.

\subsection{Database}

Our database consists of data from 5 actual operating trucks normally driving around Europe, provided by Volvo Group, participating in this research.
Three of the trucks are equipped with a regular physical mirror, and the two others with a mirrorless CMS (Camera Monitor System).
The trucks are equipped with a data logger that captures CAN signals at 10 Hz.
The common signals available in all trucks and employed for this work are listed below:

\begin{enumerate}
    \item Position of the accelerator pedal
    \item Distance to the vehicle ahead
    \item Speed of the vehicle ahead
    \item Relative speed difference between the vehicle and the left wheel 
    \item Vehicle speed
    \item Vehicle lateral acceleration
    \item Vehicle longitudinal acceleration
    \item Lane change status of the vehicle
    \item Status of the left turn indicator
    \item Status of the right turn indicator      
\end{enumerate}

Signals 1 to 7 vary continuously within specific ranges, whereas signals 8 to 10 have discrete statuses related to activation/no activation.
Other CAN signals are also available in the data logger, but they are not considered in this work since we consider that they are not informative of an overtake manoeuvre. These include, for example, the position of the brake pedal, the position of the gear lever, or the status and speed set in the cruise control.

To avoid storage issues and to obtain segments where a potential overtake occurs, we defined a precondition trigger so that the data logger only records when it is met. 
Such trigger is activated based on specific thresholds to the following signals: signal 8 (active), signal 5 (more than 50 km/h), signal 2 (less than 200 m), and signal 4 (more than 0.1 km/h).
If the trigger is activated, the logger saves the CAN signals from 20 seconds before the trigger up to 45 seconds thereafter. 
The logger also saves video from a camera placed on the dashboard looking ahead of the vehicle.
The files are later labelled manually by watching the videos and determining if the recorded segment is an overtake or not.
Afterwards, the video data is discarded for further use.

\begin{table}[]
\centering
\caption{Files available per truck and per class. tx denotes the truck number (given by x). class0=no-overtake. class1=overtake.}
\label{tab:db-files}
\begin{tabular}{llll|l}

\multicolumn{5}{c}{} \\

Type                    & Truck & class0 & class1 & total \\ \hline \hline
                        & t1    & 125    & 417    & 542   \\
      \textit{Mirror}            & t2    & 163    & 83     & 246   \\
                        & t3    & 8      & 11     & 19             \\ \cline{2-5}
                        & total & 296    & 511    & 807   \\ \hline \hline
                        & t4    & 81     & 342    & 423   \\
       \textit{CMS}              & t5    & 5      & 12     & 17    \\ \cline{2-5}
                        & total & 86     & 354    & 440  \\ \hline \hline
Both                    & total & 382    & 865    & 1247   \\ \hline                           \hline
\end{tabular}%
\end{table}

After this process, we obtained the files indicated in Table~\ref{tab:db-files}. 
Compared to our previous work \citep{Butt2024}, we have increased the number of files of trucks t1-t3 from 712 to 807.
Also, we incorporate data from trucks t4-t5, which were not available in the previous paper.
The precondition trigger is designed to detect when the vehicle is to change lane (signal 8), to be sufficiently close to the vehicle ahead (signal 2), and to move laterally to the left (signal 4), which are indicative signs of a potential overtake. 
However, around 31\% of the obtained files correspond to other driving situations where the truck can show similar dynamics. After watching the videos, it occurs, for example, when turning left at an intersection, changing lanes to leave an exit-only lane, giving way to a vehicle merging into the road, or surpassing a stopped vehicle or construction workers.
The left turn indicator (signal 9) could be another possible precondition indicator, but many other driving situations, such as the ones mentioned, could also lead to false positives.
It has also been acknowledged that the use of such indicators can be low, so while they can be helpful, they are not sufficient for manoeuvre prediction \citep{khairdoost2020real}.
On the other hand, the minimum speed condition (signal 5) is designed to filter out city traffic events, leaving the same footprint in signals 8, 2 and 4, but that are not really overtakes.
Since the trucks were operating freely in real traffic, it was not possible to enforce strict control over traffic or environmental conditions during data collection. However, the triggering rules used help ensure that the selected periods correspond to active driving, reducing the inclusion of idle, city or stop-and-go scenarios, and providing files mostly from highways or non-urban roads.
%

\subsection{Machine Learning Methods}

We compare three machine learning methods in this work: Artificial Neural Networks (ANN) \citep{haykin2009neural}, Random Forest (RF) \citep{[Breiman01rf]}, and Support Vector Machines (SVM, with linear and rbf kernels) \citep{[Vapnik95]}. 
These are based on different classification strategies and are a popular choice in the related literature on detecting vehicle manoeuvres, as seen in Table~\ref{tab:summary-related-works} \citep{kim2017prediction,das2020detecting,das2023lanechange} and other previous surveys \citep{xing2019driver}:

\begin{itemize}
    \item \textbf{ANN} consists of several interconnected neurons arranged in layers (i.e., input, hidden, and output layers). The nodes in one layer are interconnected to all nodes in the neighbouring (previous and subsequent) layers. Two design parameters of ANNs are the number of intermediate layers and the number of neurons per layer.

    \item \textbf{RF} is an extension of the standard classification trees algorithm. It is an ensemble method where the results of many decision trees are combined. Such a combination helps to reduce overfitting and to improve generalisation capabilities. The trees in the ensemble are grown by using bootstrap samples of the data. 

    \item \textbf{SVM} searches for an optimal hyperplane in a high-dimensional space that separates the data into two classes. Different kernel functions can be used to transform data that can be used to form the hyperplane, such as linear, or gaussian (rbf).
\end{itemize}

\subsection{Experimental Protocol}

We crop the files around the precondition trigger and use only the samples that fall within the defined crop. We crop all possible combinations of -20 seconds, -10 seconds and -5 seconds before the trigger (which we call \textit{starttrigger}), and +0 seconds, +1 second, +2 seconds and +5 seconds after the trigger (\textit{endtrigger}).   
The size of the crop (also called \textit{time window}) has been the subject of discussion in the literature \citep{xing2019driver,khairdoost2020real}. Since the existing literature has employed different crop sizes around the event of interest, here we consider several possibilities to assess the impact of its size. 
Afterwards, the CAN signals are processed with a sliding window of size $w$. Again, the many possibilities in the literature lead us to consider different sliding window sizes, in particular $w$ = 0, 0.5, 1 and 2 seconds, with 50\% overlap. 
Table~\ref{tab:db-samples-per-file} gives the number of samples per crop considering the different combinations of \textit{starttrigger}, \textit{endtrigger} and sliding window sizes.
For signals 1-7 (non-categorical), we compute the mean and standard deviation of the samples inside the sliding window \citep{kim2017prediction}, whereas for signals 8-10 (categorical), we extract the majority value among the samples.
%
%
Obviously, for a sliding window size of $w$ = 0, we take the raw samples of the file directly, without any mean, standard deviation or majority computation, thus keeping the 10 CAN data channels.
When $w \neq$ 0, we effectively duplicate the number of non-categorical signals from 7 to 14, resulting in 17 CAN channels available.

\begin{table}[]
\centering
\caption{Available samples per file and CAN channels given the different possibilities for cropping the files around the precondition trigger and the sliding window size $w$. Find more details in the text. 
}
\label{tab:db-samples-per-file}
\resizebox{\textwidth}{!}{%
\begin{tabular}{l|llll|llll|llll|llll|}

\multicolumn{17}{c}{} \\
\cline{2-17}
 
 & \multicolumn{16}{c|}{\textbf{\textit{endtrigger} (seconds)}} \\ 
 
 & \multicolumn{4}{c}{\textbf{0}} & \multicolumn{4}{c}{\textbf{1}} & \multicolumn{4}{c}{\textbf{2}} & \multicolumn{4}{c|}{\textbf{5}} \\ 
 \hhline{~================}
 
 & \multicolumn{4}{c|}{\textbf{\begin{tabular}[c]{@{}c@{}}sliding window $w$\\ (seconds)\end{tabular}}} & \multicolumn{4}{c|}{\textbf{\begin{tabular}[c]{@{}c@{}}sliding window $w$\\ (seconds)\end{tabular}}} & \multicolumn{4}{c|}{\textbf{\begin{tabular}[c]{@{}c@{}}sliding window $w$\\ (seconds)\end{tabular}}} & \multicolumn{4}{c|}{\textbf{\begin{tabular}[c]{@{}c@{}}sliding window $w$\\ (seconds)\end{tabular}}} \\ 
 
\textbf{\begin{tabular}[c]{@{}l@{}}\textit{starttrigger}\\ (seconds)\end{tabular}} & \textbf{0} & \textbf{0.5} & \textbf{1} & \textbf{2} & \textbf{0} & \textbf{0.5} & \textbf{1} & \textbf{2} & \textbf{0} & \textbf{0.5} & \textbf{1} & \textbf{2} & \textbf{0} & \textbf{0.5} & \textbf{1} & \textbf{2} \\ \hline

\textbf{-20} & 201 & 66 & 39 & 19 & 211 & 69 & 41 & 20 & 221 & 72 & 43 & 21 & 251 & 82 & 49 & 24 \\
\textbf{-10} & 101 & 32 & 19 & 9 & 111 & 36 & 21 & 10 & 121 & 39 & 23 & 11 & 151 & 49 & 29 & 14 \\
\textbf{-5} & 51 & 16 & 9 & 4 & 61 & 19 & 11 & 5 & 71 & 22 & 13 & 6 & 101 & 32 & 19 & 9 \\ \hline

CAN channels & 10 & 17 & 17 & 17 & 10 & 17 & 17 & 17 & 10 & 17 & 17 & 17 & 10 & 17 & 17 & 17 \\ \hline

\end{tabular}%
}
\end{table}

All samples from overtake files are then labelled as class1 (positive class or overtake), whereas all samples for no-overtake files are labelled as class0 (negative class or no-overtake).
The classifiers are trained with a portion of the files from the trucks. 
All other available files are used for testing. 
The exact number of training and test files per truck is given in Table~\ref{tab:db-training-test-files}.
To avoid bias due to class imbalance during training, we ensure that the training dataset is balanced, containing the same number of files per truck and per class.
%
%
To do so, we check how many files of each class are available per truck, then we take 70\% of the minimum.
The test set is imbalanced, which we address by reporting results using AUC-ROC (Area Under the Receiver Operating Characteristic Curve), which is robust to class imbalance, and by providing TPR and TNR separately rather than using global accuracy.
Experiments are conducted using Matlab r2023b. All classifiers are left with the default values (ANN: one hidden layer with 10 neurons; RF: 100 decision trees), except:

\begin{itemize}
    \item ANN and SVM: we use standardisation in non-categorical signals by subtracting the mean and dividing by the standard deviation of the training samples
    \item The ANN iteration limit is raised to 1e6 (from 1e3) and the SVMrbf iteration limit is raised to 1e8 (from 1e6) to facilitate convergence
\end{itemize}

In addition, all classifiers are set up to provide class posterior probabilities as output. Thus, their output is a continuous value in $[0,1]$. If the output is higher than 0.5, the input is classified to belong to the positive class, and to the negative class if lower than 0.5.

\begin{table}[]
\centering
\caption{Files employed for training and test per class and per truck given the different possibilities for cropping the files around the precondition trigger.}
\label{tab:db-training-test-files}
\resizebox{0.85\textwidth}{!}{%
\begin{tabular}{cl|llll|llll|llll|}

\multicolumn{14}{c}{} \\
 \cline{3-14}

\multicolumn{1}{l}{} &  & \multicolumn{4}{c|}{\textbf{\begin{tabular}[c]{@{}c@{}}training\\ (both classes)\end{tabular}}} & \multicolumn{4}{c|}{\textbf{\begin{tabular}[c]{@{}c@{}}test class0\\(no overtake)\end{tabular}}} & \multicolumn{4}{c|}{\textbf{\begin{tabular}[c]{@{}c@{}}test class1\\(overtake)\end{tabular}}}  \\ 
\hhline{~~============}
 
\multicolumn{1}{l}{} &  & \multicolumn{4}{c|}{\textbf{\begin{tabular}[c]{@{}c@{}}\textit{endtrigger} \end{tabular}}} & \multicolumn{4}{c|}{\textbf{\begin{tabular}[c]{@{}c@{}}\textit{endtrigger}\end{tabular}}} & \multicolumn{4}{c|}{\textbf{\begin{tabular}[c]{@{}c@{}}\textit{endtrigger}\end{tabular}}}  \\

\multicolumn{1}{l}{\textbf{truck}} & \textbf{\begin{tabular}[c]{@{}l@{}}\textit{starttrigger}\\ \end{tabular}} & \textbf{0} & \textbf{1} & \textbf{2} & \textbf{5} & \textbf{0} & \textbf{0.5} & \textbf{1} & \textbf{2} & \textbf{0} & \textbf{0.5} & \textbf{1} & \textbf{2} \\ \hline

\textbf{t1} & \textbf{-20} & 72 & 72 & 72 & 68 & 35 & 35 & 34 & 33 & 325 & 325 & 324 & 328\\
 & \textbf{-10} & 74 & 74 & 74 & 70 &  36 & 36 & 35 & 34 & 336 & 336 & 335 & 339 \\
 & \textbf{-5} & 79 & 79 & 78 & 74 &  38 & 37 & 37 & 35 & 333 & 333 & 333 & 337 \\ \hline

\textbf{t2} & \textbf{-20} & 37 & 37 & 37 & 37  &  117 & 117 & 117 & 117 & 43 & 43 & 43 & 43 \\
 & \textbf{-10} & 38 & 38 & 38 & 38 &  119 & 119 & 119 & 119 & 45 & 45 & 45 & 45\\
 & \textbf{-5} & 38 & 38 & 38 & 38 &  121 & 121 & 121 & 121  & 45 & 45 & 45 & 45\\ \hline
 
\textbf{t3} & \textbf{-20} & 4 & 4 & 4 & 4 & 2 & 2 & 2 & 2 & 7 & 7 & 7 & 7 \\
 & \textbf{-10} & 4 & 4 & 4 & 4 & 2 & 2 & 2 & 2 & 7 & 7 & 7 & 7\\
 & \textbf{-5} & 4 & 4 & 4 & 4  & 3 & 3 & 3 & 3 & 7 & 7 & 7 & 7 \\ \hline

\textbf{t4} & \textbf{-20} & 46 & 45 & 44 & 42 & 21 & 20 & 19 & 18 & 259 & 260 & 261 & 263\\
 & \textbf{-10} & 50 & 49 & 47 & 45 & 22 & 21 & 21 & 20 & 270 & 271 & 273 & 275\\
 & \textbf{-5} & 52 & 51 & 49 & 47 &  23 & 22 & 22 & 21 & 282 & 283 & 285 & 287  \\ \hline

\textbf{t5} & \textbf{-20} & 3 & 3 & 3 & 3 &  2 & 2 & 2 & 2 & 8 & 8 & 8 & 8\\
 & \textbf{-10} & 3 & 3 & 3 & 3 & 2 & 2 & 2 & 2 & 8 & 8 & 8 & 8\\
 & \textbf{-5} & 3 & 3 & 3 & 3 & 2 & 2 & 2 & 2 & 8 & 8 & 8 & 8\\ \hline
 
\end{tabular}%
}
\end{table}

\section{Results}
\label{sect:results}

To keep consistency with our previous work \citep{Butt2024}, we start the analysis only with trucks t1-t3. This is because data of these trucks was the only available data when this research started, in particular when we initially trained the classifiers with different combinations of \textit{starttrigger}, \textit{endtrigger} and sliding window sizes.
Later on, when data from trucks t4-t5 was made available, we tested the classifiers trained with t1-t3 and also considered retraining the classifiers with data from all trucks.

\begin{figure} [htb]
\centering
\includegraphics[width=0.95\textwidth]{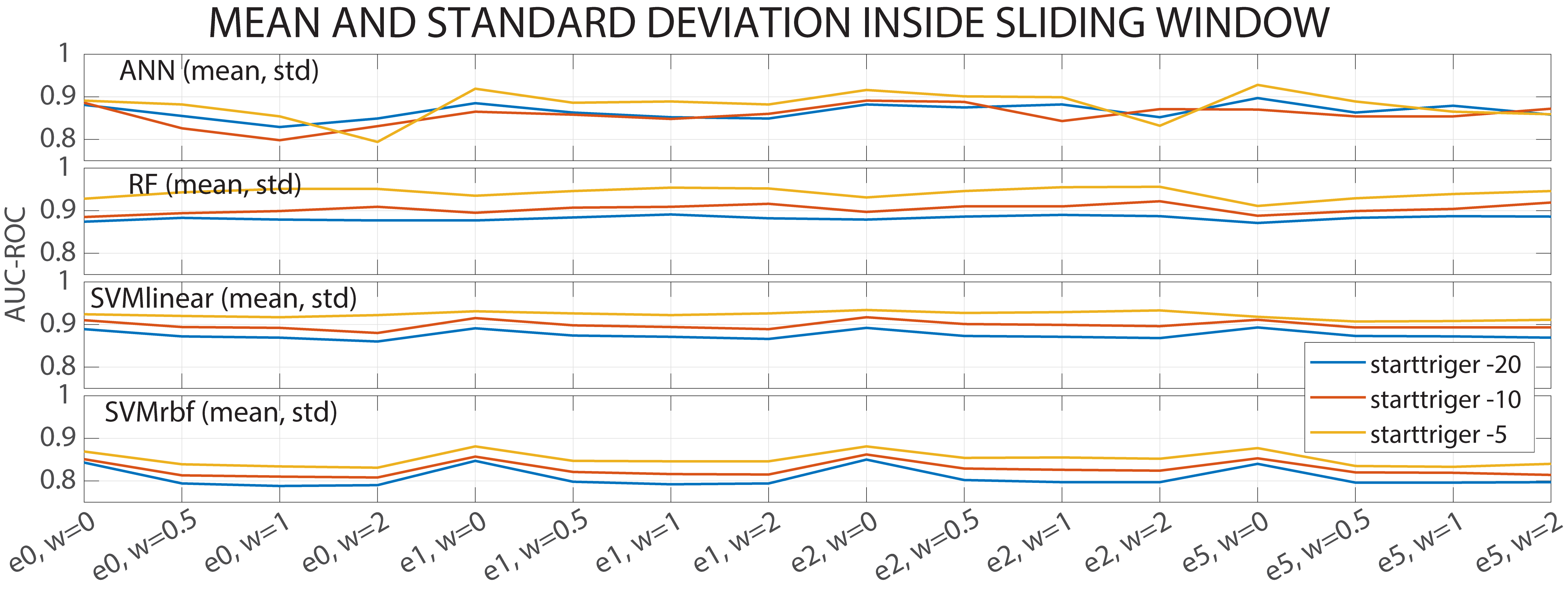}
\includegraphics[width=0.95\textwidth]{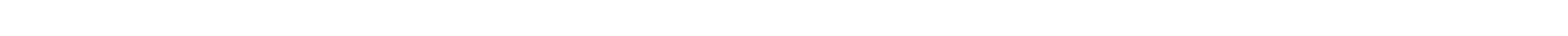}
\includegraphics[width=0.95\textwidth]{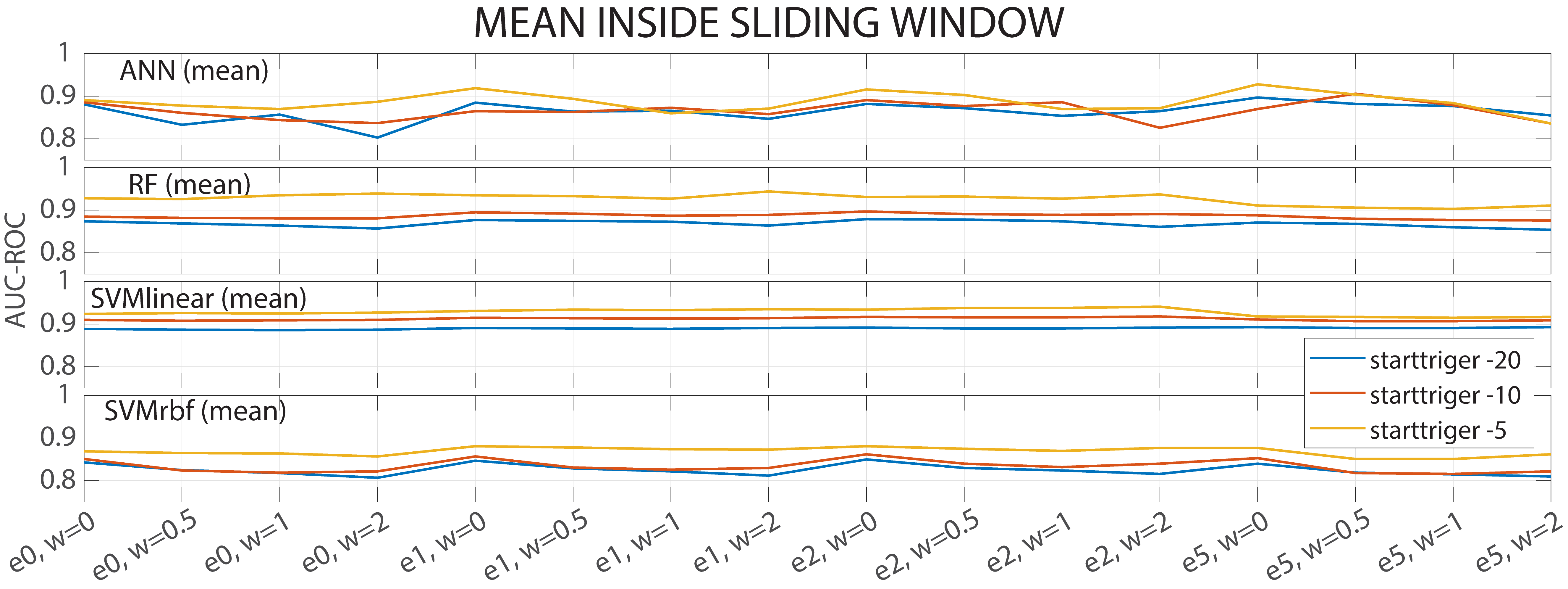}
\caption{AUC-ROC values for trucks t1-t3 (\textit{mirror} trucks) when the classifiers are trained with data from trucks t1-t3. Results are given for different combinations of crops before the precondition trigger (\textit{starttriger} in the legends), after the precondition trigger (parameter \textit{e} in the horizontal axes) and sliding window sizes (parameter \textit{w}). Better seen in colour.}
\label{fig:AUC_ROC}
\end{figure}

\subsection{Parameter analysis of the classifiers}

We first analyse (Figure~\ref{fig:AUC_ROC}) the performance of \textit{mirror} trucks (t1-t3) in terms of the AUC-ROC of the classifiers, considering the values of \textit{starttrigger}, \textit{endtrigger}, and sliding window sizes tested in this study. 
We also distinguish between computing both the mean and standard deviation of non-categorical signals within the sliding window (top) or only the mean (bottom).
The AUC-ROC 
summarises the ability of a model to discriminate between positive and negative instances across all classification thresholds. The AUC-ROC score ranges from 0 to 1, where 0.5 indicates random guessing, and 1 indicates perfect classification performance. 
As can be observed, the AUC-ROC is inversely correlated in the majority of cases with the length of \textit{starttrigger}, i.e. the closer to the actual overtaking event, the better the accuracy. 
This is reflected by the orange curve (\textit{starttrigger}=-5) being above the red and blue ones (\textit{starttrigger}=-10/-20).
A larger \textit{starttrigger} crop may introduce unnecessary data too early in time that can confuse the models, reducing their ability to discriminate between overtaking and non-overtaking scenarios.
On the other hand, this highlights the potential challenge of detecting the event of interest well in advance of its onset.

\begin{figure} [htb]
\centering
\includegraphics[width=0.9\textwidth]{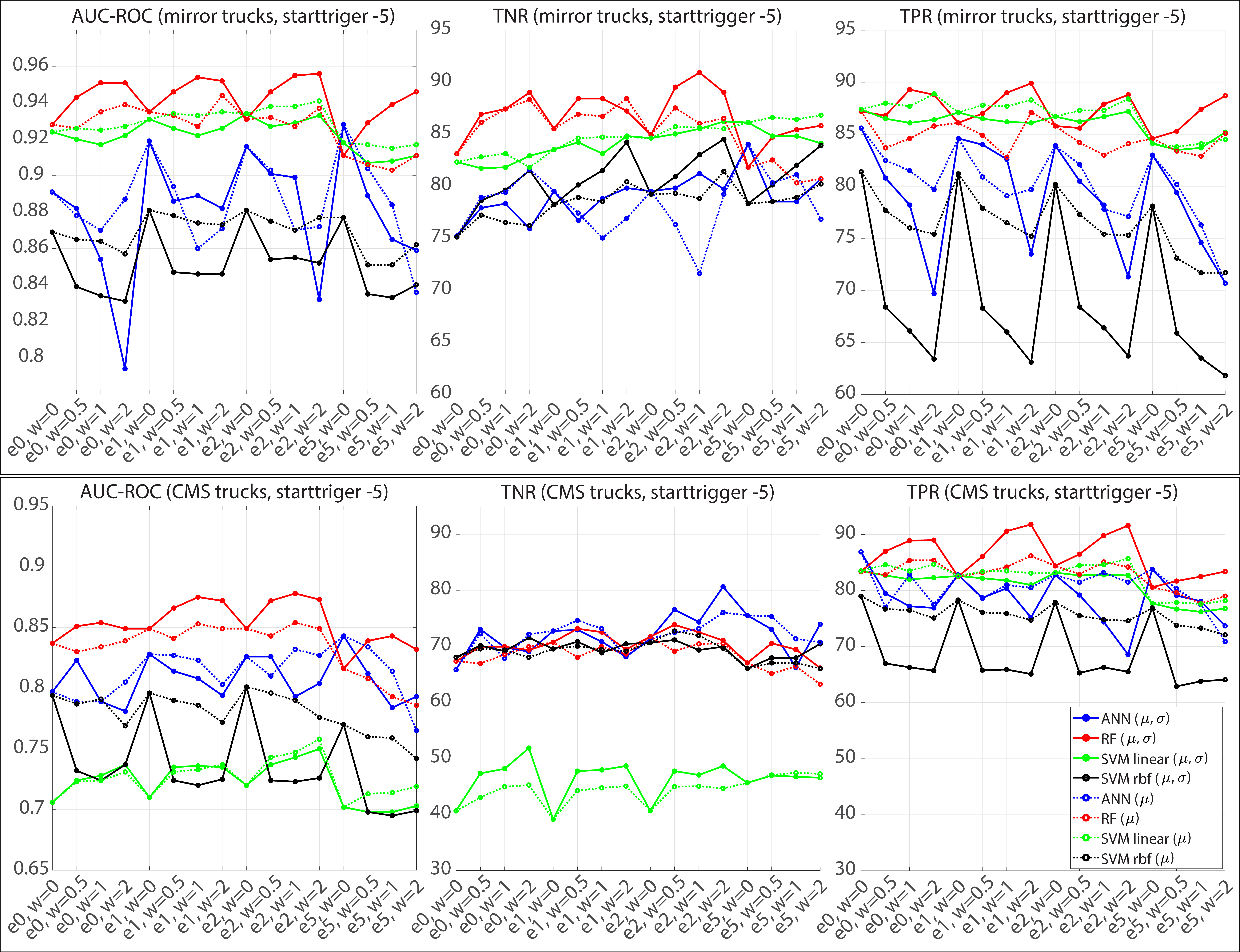}
\caption{AUC-ROC, \textit{TNR} and \textit{TPR} values for trucks t1-t3 (top, \textit{mirror} trucks) and t4-t5 (bottom, \textit{CMS} trucks) when the classifiers are trained with data from trucks t1-t3. The plot uses a value of \textit{starttriger}=-5 (the best case of Figure~\ref{fig:AUC_ROC}). Solid lines indicate that the mean and standard deviation of non-categorical signals are computed inside the sliding window, whereas dashed lines indicate that only the mean is computed. ex in the x-axes denotes the length of \textit{endtrigger} (given by x), and $w$ denotes the sliding window size. Better seen in colour.}
\label{fig:AUC_ROC_TNR_TPR_start_minus_5}
\end{figure}

Given the superior performance of \textit{starttrigger}=-5, we focus on this configuration and include as metrics the True Positive Rate (\textit{TPR}) and True Negative Rate (\textit{TNR}). The \textit{TPR} measures the proportion of positive instances (overtakes) correctly classified as positives, while the \textit{TNR} (also referred to as specificity) indicates the proportion of negative instances (no-overtakes) correctly classified as negatives.
\textit{TPR} is calculated as \textit{TPR} = \textit{TP} / (\textit{TP} + \textit{FN}), where \textit{TP} is the number of true positives, and \textit{FN} is the number of false negatives. 
Similarly, \textit{TNR} is calculated as \textit{TNR} = \textit{TN} / (\textit{TN} + \textit{FP}), where \textit{TN} is the number of true negatives, and \textit{FP} is the number of false positives.
Figure~\ref{fig:AUC_ROC_TNR_TPR_start_minus_5} (top row) presents the results for the \textit{mirror} trucks.

The first observation is that both \textit{TNR} and \textit{TPR} appear relatively balanced across all classifiers. That is, for each classifier, both metrics are generally within similar ranges, with only slight variations. 
When examining \textit{TNR}, Random Forest (red curves) outperforms the other classifiers, followed by SVM linear (green). In contrast, ANN (blue) and SVM rbf (black) are the worst, with their \textit{TNR} in similar ranges.
For \textit{TPR}, Random Forest and SVM linear show comparable accuracy, followed by ANN and then SVM rbf. 
%
%
%
Another observation is that using only the mean (dashed curves) vs. both the mean and standard deviation (solid curves) is mostly irrelevant for \textit{TNR}, but \textit{TPR} is more sensitive. The solid curves, which incorporate both the mean and standard deviation, have greater \textit{TPR} variability, being dependent on the size of the sliding window.
In addition, which option is the best depends on the classifier. RF performs better when using both the mean and standard deviation (solid red curve), whereas the other classifiers prefer the mean only (dashed curves).

Interestingly, the optimal sliding window size varies between classifiers too. Both ANN and SVMs perform best with \textit{w}=0 (no sliding window, or using the raw samples), making the discussion of mean vs. both mean and standard deviation irrelevant. On the other hand, Random Forest shows the best performance with larger window sizes, specifically \textit{w}=1 or 2.
Regarding the impact of \textit{endtrigger} (the end of the crop after the precondition trigger), we observe that \textit{TNR} benefits from a larger crop, with curves increasing towards the right of the \textit{x}-axis. 
A larger crop provides more context and allows the model to confirm that no overtaking is taking place but another manoeuvre instead, reducing false positives. 
Conversely, \textit{TPR} has an opposite trend, suggesting that the most useful information for detecting the overtake class disappears shortly after the trigger.
As we will see later (Section~\ref{sect:can_signals_analysis} and second column of Figures~\ref{fig:boxplots_mirror}-\ref{fig:boxplots_CMS}), the dynamics of the CAN signals are different before and after an overtake starts. 
As a result, 
%
extending the crop too far beyond the trigger introduces unnecessary data that can lead to an increase in false negatives.

\subsection{Comparison of \textit{mirror} and \textit{CMS} trucks}

We then analyse the performance of the classifiers on \textit{CMS} trucks (t4-t5). Results are given in the bottom row of Figure~\ref{fig:AUC_ROC_TNR_TPR_start_minus_5}.
Recall that the algorithms are still trained on data from \textit{mirror} trucks (t1-t3) only, so in this sub-section, we are evaluating how the classifiers behave when tested on trucks that differ from the training set.
%

Comparing the top and bottom rows of Figure~\ref{fig:AUC_ROC_TNR_TPR_start_minus_5}, we observe that \textit{TPR} remains consistent across truck types for all classifiers. In some cases, it even improves, e.g. the best performance of RF exceeds 90\%, or the variability of ANN and SVM rbf with respect to \textit{w} is less pronounced. 
These findings suggest that the `execution' of an overtake remains largely invariant, so the \textit{TPR} is not significantly impacted when trained and tested on overtakes from different trucks or drivers. 
However, \textit{TNR} shows a significant decline in \textit{CMS} trucks, which significantly affects the AUC-ROC. The drop is particularly pronounced for SVM linear, which goes from having the second-best \textit{TNR} to performing worse than a random guess.
This suggests that the patterns of no-overtake signals differ between mirror and CMS datasets, a phenomenon that we will further analyse in Section~\ref{sect:can_signals_analysis}.

The majority of observations regarding the optimal sliding window size or the impact of \textit{endtrigger} also hold in this sub-section. 
ANN and SVMs perform best with \textit{w}=0 (at least in terms of \textit{TPR}), while RF does better with larger windows (\textit{w}=1 or 2).
Similarly, \textit{TPR} worsens if the end crop is extended too far beyond the precondition trigger. 
Additionally, RF prefers the use of mean and standard deviation as features, whereas the other classifiers perform better with the mean only.

\begin{figure} [htb]
\centering
\includegraphics[width=0.73\textwidth]{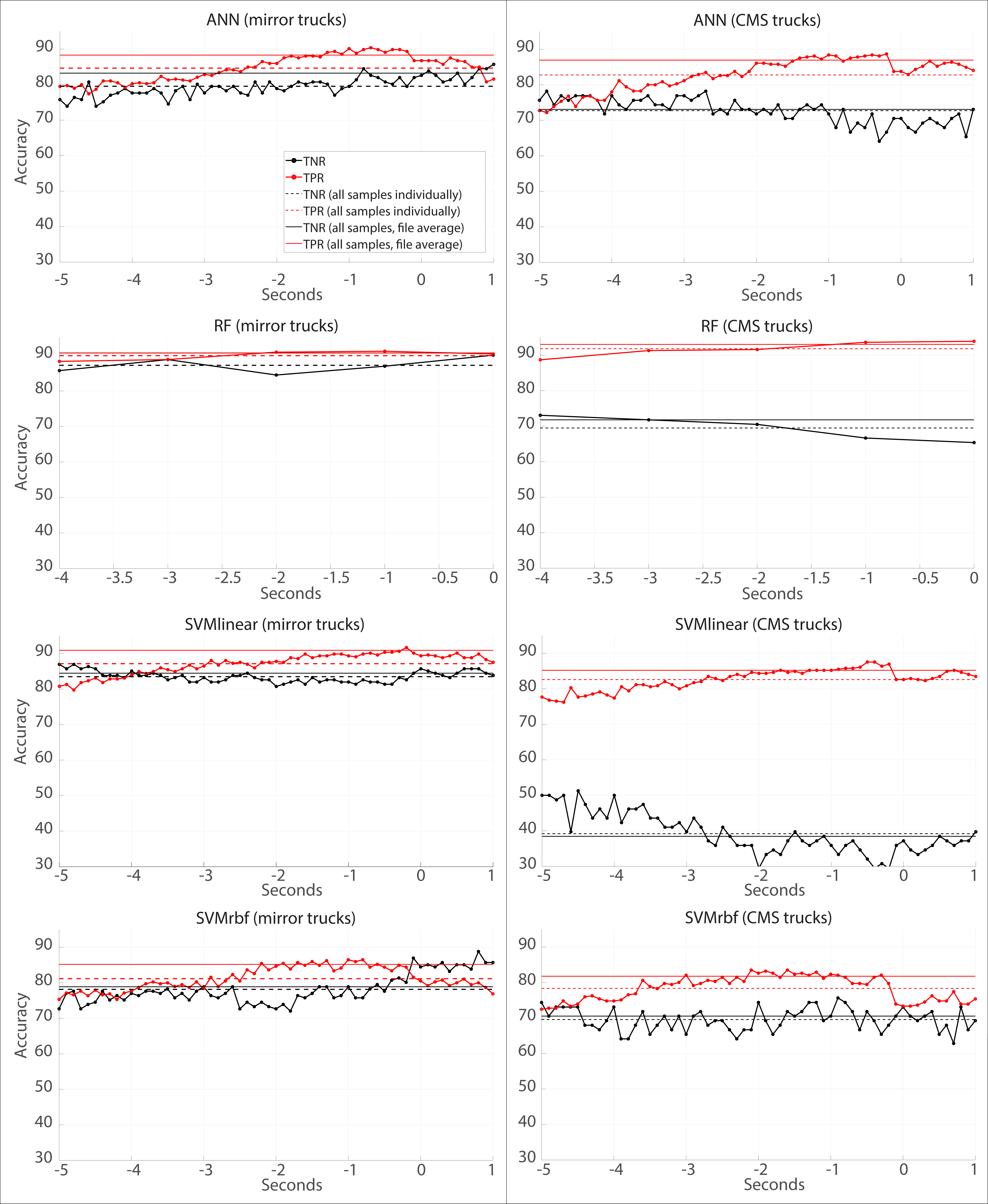}
\caption{TNR/TPR of the classifiers at different moments before/after the precondition trigger for trucks t1-t3 (left, \textit{mirror} trucks) and t4-t5 (right, \textit{CMS} trucks) when the classifiers are trained with data from t1-t3. See Section~\ref{sect:prediction_window_analysis} for more details about the configuration of the experiments. Better seen in colour.
}
\label{fig:TNR_TPR_per_time}
\end{figure}

\subsection{Prediction window analysis}
\label{sect:prediction_window_analysis}

From the results of previous sub-sections, we carry forward the following best configurations for further analysis: \textit{starttrigger}=-5, \textit{endtrigger}=1 (a compromise between a larger crop for higher \textit{TNR} vs. a shorter crop for higher \textit{TPR}), and \textit{w}=0 for all classifiers (except \textit{w}=2 for RF, which is applied using mean and standard deviation within the sliding window).
We then analyse in this sub-section the \textit{TNR}/\textit{TPR} at different time points around the precondition trigger (Figure~\ref{fig:TNR_TPR_per_time}).
To obtain these plots, \textit{TNR}/\textit{TPR} are computed using only the samples corresponding to each specific time point. The classifiers remain trained exclusively on data from t1-t3.
The following additional results are also shown in the figure:

\begin{enumerate}
    \item Per-sample \textit{TNR}/\textit{TPR} (dashed horizontal lines), computed using all samples from each file individually. This generates one decision score per sample of the file, corresponding to how accuracy metrics were previously calculated for Figures~\ref{fig:AUC_ROC} and \ref{fig:AUC_ROC_TNR_TPR_start_minus_5}.
    
    \item Per-file \textit{TNR}/\textit{TPR} (solid horizontal lines), computed by averaging the decision scores of all samples within a file. This produces one decision score per file, integrating the output of the classifiers for all samples in the time crop.
    
\end{enumerate}

With \textit{mirror} trucks (t1-t3), left column, the accuracy (both \textit{TNR} and \textit{TPR}) is observed to improve as we approach the trigger. 
After the trigger at $x$=0, \textit{TPR} (red curves) decreases. This is consistent with the previous observation that a large \textit{endtrigger} value was detrimental to \textit{TPR}. The findings of this subsection confirm that the most relevant information for the overtake class is contained before the trigger.
The curves also confirm the previous observation that \textit{TNR}/\textit{TPR}
are balanced with these trucks, i.e. they are in the same range for most classifiers, with differences of 5-10\% maximum at some time points.

It can also be seen that the solid horizontal lines (integrating the scores of the entire crop) are a good option to obtain high accuracy, counteracting the variations of accuracy with time. 
The downside is that it needs to wait until all samples of the crop are available to make a decision. 
However, with a classifier like RF, both \textit{TNR} and \textit{TPR} are close to or higher than 90\% at all times, highlighting the power of this classifier for our task.

With \textit{CMS} trucks (t4-t5), right column, the same trend is observed for \textit{TPR}, i.e. it increases as we approach the trigger. 
However, \textit{TNR} actually decreases towards the trigger. 
Also, as observed in the previous sub-section, \textit{TNR} with \textit{CMS} trucks is much lower than \textit{TPR}.
We attribute this to a difference in the patterns of no-overtake between the two trucks' sets, which we will analyse in the following sub-section.



\begin{figure} [H]
\centering
\includegraphics[width=0.75\textwidth]{3-A4_boxplots_MIRROR.png}
\caption{Mirror trucks (t1-t3): boxplots of the CAN signals $\pm$20 seconds around the precondition trigger. Left column: no-overtake class. Right: overtake class. Better seen in colour.}
\label{fig:boxplots_mirror}
\end{figure}

\begin{figure} [H]
\centering
\includegraphics[width=0.75\textwidth]{3-A4_boxplots_CMS.png}
\caption{CMS trucks (t4-t5): boxplots of the CAN signals $\pm$20 seconds around the precondition trigger. Left column: no-overtake class. Right: overtake class. Better seen in colour.}
\label{fig:boxplots_CMS}
\end{figure}


\subsection{CAN signals analysis}
\label{sect:can_signals_analysis}

In this sub-section, we directly examine the patterns of the CAN signals to understand the dynamics and behaviour of the drivers in an overtake.
It will also provide insights into the observed performance differences between t1-t3 (\textit{mirror} trucks) and t4-t5 (\textit{CMS} trucks).
We do the analysis by plotting in Figures~\ref{fig:boxplots_mirror} and \ref{fig:boxplots_CMS} the boxplots of the CAN signals of each class for the two truck groups over the interval from -20 to +20 seconds relative to the precondition trigger.

We begin by analyzing the CAN signals of \textit{mirror} trucks (t1-t3), Figure~\ref{fig:boxplots_mirror}, focusing on each signal individually:

\begin{itemize}
    \item \textit{Position of the accelerator pedal}. During overtakes, the accelerator is increasingly pushed, often exceeding 90\% as the trigger approaches. After the trigger, some drivers release the pedal slightly, but most maintain high pressure. 
    In no-overtake scenarios, the pedal position ranges from not pushed to moderately pushed, likely indicating the use of cruise control in many cases, since the median is always at 0.  
    
    \item \textit{Distance to the vehicle ahead}. In overtakes, the median distance steadily decreases below 50m before abruptly jumping to 255m after the trigger, signalling a lane change (no vehicle ahead). 
    Occasionally, other vehicles are detected beyond 100m after the trigger, suggesting that the new lane is not empty. 
    The majority of lane changes occur around 25m from the vehicle ahead, though some happen at greater distances. 
    In no-overtakes, the median distance remains consistently high, around 100-125m, with occasional values at 255m, indicating no vehicle ahead.
    
    \item \textit{Speed of the vehicle ahead}. This signal behaves in a similar way to the previous one, with minor differences. 
    In overtakes, the vehicle ahead maintains a constant median speed, typically around 90 km/h before the trigger, consistent with either another truck (speed-limited by software) or a road limit. 
    After the trigger, the signal then jumps abruptly.
    Segments with a median of 120 km/h likely correspond to motorways.    
    
    \item \textit{Relative speed vehicle/wheel}. In overtakes, this signal activates primarily around the trigger. It initially takes negative values (wheel turned left), then rises to positive values (wheel turned right, returning to straight), and eventually stabilises at zero (truck moving straight). 
    In no-overtake scenarios, the signal is mostly zero or positive, indicative of other manoeuvres.
        
    \item \textit{Vehicle speed}. During overtakes, the median speed remains close to 90 km/h, with minimal variability, indicating the truck is near its maximum speed before overtaking. 
    In no-overtakes, speed is more variable, with a lower median and a slight increase over time.
    
    \item \textit{Vehicle lateral acceleration}. In both classes, lateral acceleration is positive due to the leftward motion that triggers the precondition. However, in overtakes, a sharper acceleration increase is observed at the trigger, followed by a pronounced decrease or even negative values afterwards. This suggests a lane change, distinguishing it from other manoeuvres like turning left.
    
    \item \textit{Vehicle longitudinal acceleration}. In overtakes, longitudinal acceleration is minimal, as the truck is already moving at high speed in most cases. A few instances show slight acceleration at the trigger, very likely those which were seen to increase speed after the trigger. 
    At the end of an overtake, a deceleration phase (negative values) might be expected if the truck slows down to adapt its speed when returning to the right lane. However, the absence of such deceleration in the plot suggests that, in the analysed segments, overtakes extend beyond the 20-second time window shown in the $x$-axis.
    In no-overtakes, the observed gradual speed increase explains the consistently positive acceleration values.
    
\end{itemize}

\begin{figure} [htb]
\centering
\includegraphics[width=0.9\textwidth]{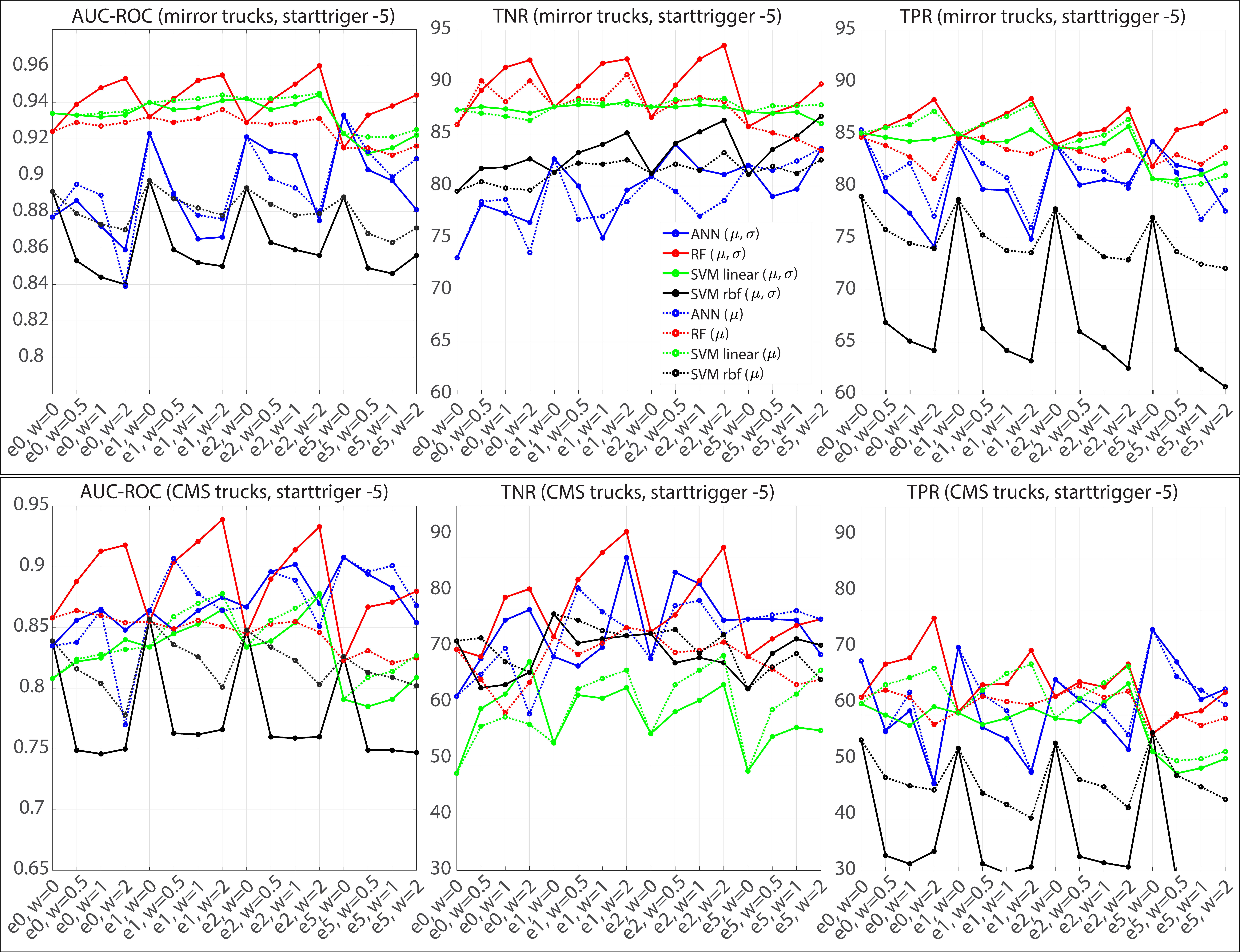}
\caption{AUC-ROC, \textit{TNR} and \textit{TPR} values for trucks t1-t3 (top, \textit{mirror} trucks) and t4-t5 (bottom, \textit{CMS} trucks) when the classifiers are trained with data from trucks t1-t5. The plot uses a value of \textit{starttriger}=-5 (the best case of Figure~\ref{fig:AUC_ROC}). Solid lines indicate that the mean and standard deviation of non-categorical signals are computed inside the sliding window, whereas dashed lines indicate that only the mean is computed. Better seen in colour.}
\label{fig:AUC_ROC_TNR_TPR_start_minus_5_train_with_CMS}
\end{figure}

These differences between classes suggest that it is feasible to distinguish them through classifier training. 
We now examine the signals of the \textit{CMS} group (\textit{t4-t5}) in Figure~\ref{fig:boxplots_CMS}, comparing them with the signals of \textit{mirror} trucks analyzed above:

\begin{itemize}
    \item \textit{Position of the accelerator pedal}. In the no-overtake class, the signal shows higher pedal positions, often exceeding 50\%, whereas \textit{mirror} trucks exhibited low values in this scenario. This indicates that in the \textit{CMS} set, drivers were not relying on cruise control as frequently during no-overtake manoeuvres, but manually managing the truck speed, so the status of the cruise control does not appear to be a reliable separator of the two classes under analysis. As argued below, no-overtake CMS data has probably been collected under heavier data traffic, not allowing for much cruise control usage.
    The overtake class, on the other hand, exhibits a consistent pattern across both truck groups.
    
    \item \textit{Distance to the vehicle ahead}. In no-overtake manoeuvres, the signal shows a lower tendency to reach 255 (indicative of no vehicle ahead). This suggests that \textit{CMS} data was more likely collected under heavier traffic conditions. 
    In contrast, \textit{mirror} trucks generally maintained higher distances or had no vehicle ahead more frequently. 
    In overtaking manoeuvres, on the other hand, both truck types exhibit similar patterns, with distances decreasing as the vehicle approaches, followed by a signal jump when the overtake is initiated, indicating no vehicle ahead.
        
    \item \textit{Speed of the vehicle ahead}. This signal reflects the same differences as observed in the distance signal, reaffirming that \textit{CMS} trucks were operating in potentially denser traffic environments.     
    
    \item \textit{Relative speed vehicle/wheel}. This signal exhibits relatively similar patterns between both classes in \textit{CMS} trucks, contrasting with the greater variability observed in mirror trucks.    
    This could reflect the random nature of the real-world driving data collection, potentially linked to conditions such as the heavier traffic noted.
    What remains consistent in overtakes is a distinct decrease towards more negative values near the trigger (indicating the wheel turned left), followed by a return to near-zero values during a few seconds (return to wheel straight).
    
    \item \textit{Vehicle speed}. The pattern of this signal is similar to the \textit{mirror} trucks, with a slightly higher median during no-overtakes. 
    
    \item \textit{Vehicle lateral acceleration}. Patterns of lateral acceleration are comparable between \textit{CMS} and \textit{mirror} trucks, indicating similar dynamics during overtakes or other lateral manoeuvres.
    
    \item \textit{Vehicle longitudinal acceleration}. While the general trend is similar to \textit{mirror} trucks, \textit{CMS} trucks exhibit less variability in acceleration during no-overtakes. 
    This aligns with the higher median speed observed earlier, reducing the need for acceleration.
    
\end{itemize}

In summary, while many signal patterns remain consistent across truck groups, key differences in no-overtake manoeuvres (particularly in speed, pedal position, and distance signals) suggest different driving contexts, possibly influenced by traffic density, transport task, and driving style variations. 
This could explain the variations in performance between the two truck groups since the classifiers were trained with data from one group only
The observed differences affect mostly the no-overtake class, which justifies the decrease in \textit{TNR} since the no-overtake patterns of the \textit{CMS} set have not been seen yet by the classifiers.
On the other hand, the behaviour of drivers during overtakes appears to be more standardised, which explains that \textit{TPR} is high and consistent across truck groups. Signals such as distance to the vehicle ahead, speed of the vehicle ahead, and lateral acceleration follow similar patterns when the trucks approach the vehicle ahead. The signals then exhibit characteristics indicative of a lane change to the left, with no vehicle ahead or at a high distance. 
%

%
%
%

\begin{figure} [htb]
\centering
\includegraphics[width=0.73\textwidth]{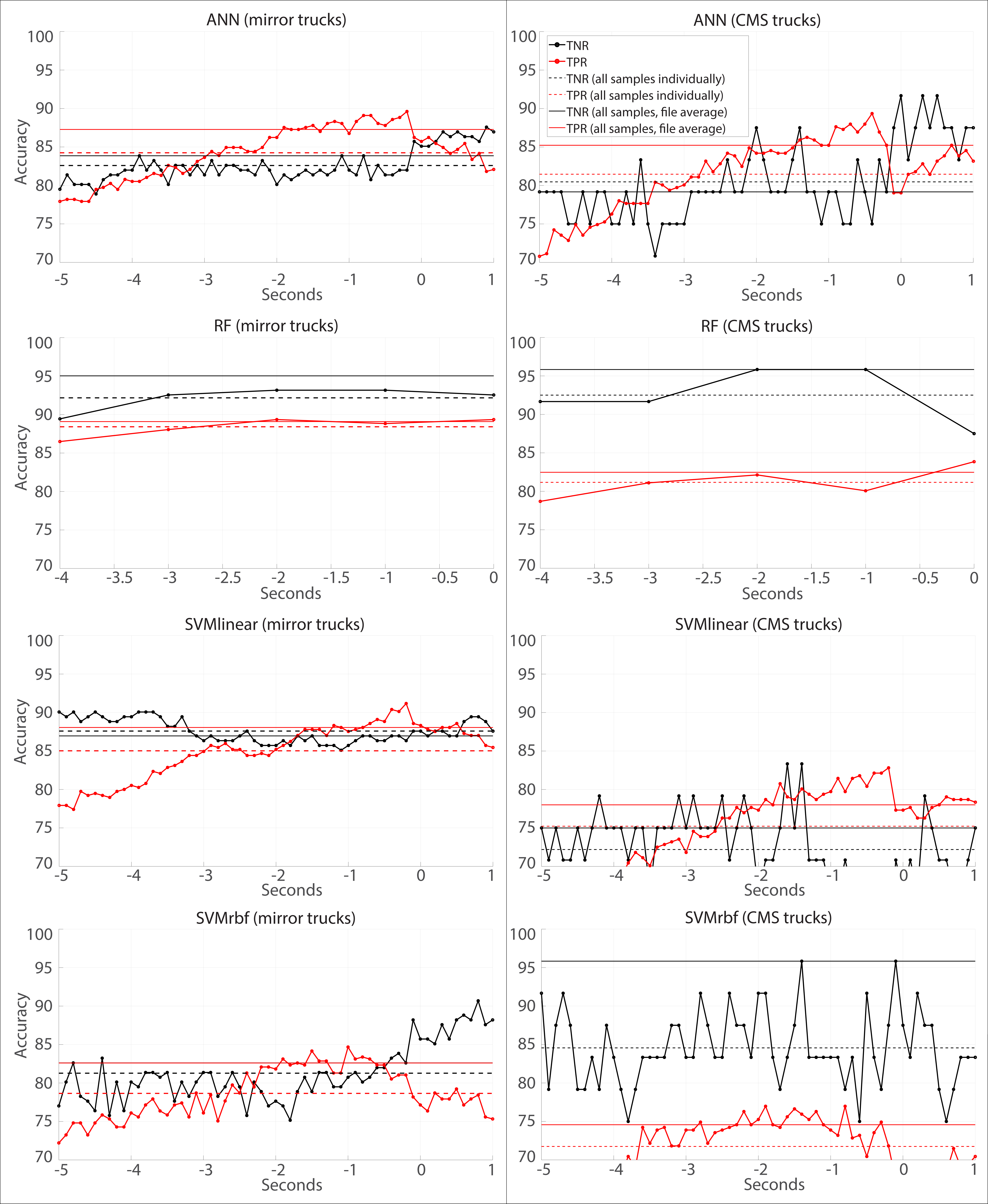}
\caption{TNR/TPR of the classifiers at different moments before/after the precondition trigger for trucks t1-t3 (left, \textit{mirror} trucks) and t4-t5 (right, \textit{CMS} trucks) when the classifiers are trained with data from t1-t5. See Section~\ref{sect:retraining} for more details about the configuration of the experiments. Better seen in colour.
}
\label{fig:TNR_TPR_per_time_train_with_CMS}
\end{figure}

\subsection{Retraining with all trucks}
\label{sect:retraining}

To counteract the observed differences in the CAN signals of the non-overtake class, we retrain the classifiers using data from all trucks. 
%
%
Figure~\ref{fig:AUC_ROC_TNR_TPR_start_minus_5_train_with_CMS} shows the \textit{AUC}, \textit{TNR} and \textit{TPR} in the same way as Figure~\ref{fig:AUC_ROC_TNR_TPR_start_minus_5}. 
It can be seen that the \textit{TNR} of both truck types generally increases when training with data from all trucks. Thus, incorporating the different observed patterns of no-overtakes improves the classifier's ability to detect them. However, \textit{TPR} is lower for both truck types. 
Since the overtake patterns appear similar between truck types based on the boxplots, the increased variability in non-overtake signals seems to impact overtake classification negatively.
One possible interpretation is that expanding the no-overtake feature space brings it closer to the overtake feature space, causing the classifiers to shift the decision boundary toward the overtake samples, thus increasing misclassification.
Other phenomena observed earlier in Figure~\ref{fig:AUC_ROC_TNR_TPR_start_minus_5} also apply here, such as the continued superiority of Random Forest, its preference for larger \textit{w} values, and the inclusion of both the mean and standard deviation of the signals. Meanwhile, other classifiers still tend to perform better with \textit{w}=0.

Figure~\ref{fig:TNR_TPR_per_time_train_with_CMS} then shows the \textit{TNR}/\textit{TPR} at different moments around the precondition trigger, in the same way as Figure~\ref{fig:TNR_TPR_per_time}.
A key observation is that now, the \textit{TNR}/\textit{TPR} of both truck types tend to remain steady or increase towards the trigger, with values mostly above 70\% in the entire crop (notice the change of range in the $y$-axes w.r.t. Figure~\ref{fig:TNR_TPR_per_time}). 
Additionally, the difference between \textit{TNR}/\textit{TPR} is now less pronounced. 
As in Figure~\ref{fig:TNR_TPR_per_time}, \textit{TPR} (red curves) decreases after the trigger, indicating that the most important information for the overtake class is located before the trigger. 

The solid horizontal lines (which integrate the scores across the entire crop) are also seen to offer an optimal solution, counteracting the oscillations of accuracy over time. 
%
%
%
%
%
%
We give in Table~\ref{tab:summary} such integrative \textit{TNR}/\textit{TPR} values depicted in Figures~\ref{fig:TNR_TPR_per_time_train_with_CMS} and \ref{fig:TNR_TPR_per_time} with solid lines. 
%
%
As observed earlier, \textit{TNR} improves for all truck types when training with data from all trucks, whereas \textit{TPR} decreases.
However, the gains in \textit{TNR} (by 25\% or more with certain classifiers) are more substantial than the decrease in \textit{TPR} (which is at most 10.5\%, and typically 2-3\%).

\begin{table}[]
\centering
\caption{\textit{TNR}/\textit{TPR} of the classifiers comparing the two training possibilities. Results are given considering a single classification score, computed as the average of scores within the crop.}
\label{tab:summary}
\resizebox{0.5\textwidth}{!}{%
\begin{tabular}{|c|l|ll|ll|}

\multicolumn{1}{l}{} & \multicolumn{1}{l}{} & \multicolumn{4}{c}{\textbf{test}}  \\ \cline{3-6}
 
\multicolumn{1}{l}{} & \multicolumn{1}{l|}{} & \multicolumn{2}{c|}{t1-t3 (\textit{mirror})} & \multicolumn{2}{c|}{t4-t5 (\textit{CMS})} \\ \hline

\textbf{classifier} & \textbf{train} & \textbf{TNR} & \textbf{TPR} & \textbf{TNR} & \textbf{TPR} \\  \hline

ANN & t1-t3 & 83.2 & \textbf{88.3} & 73.1 & \textbf{87.0} \\
    & t1-t5 & \textbf{83.9} & 87.3 & \textbf{79.2} & 85.2 \\ \hline

RF & t1-t3 & 90.7 & \textbf{90.6} & 71.8 & \textbf{93.0} \\
    & t1-t5 & \textbf{95.0} & 89.1 & \textbf{95.8} & 82.5 \\ \hline

SVM linear & t1-t3 & 84.5 & \textbf{90.9} & 38.5 & \textbf{85.2} \\
    & t1-t5 & \textbf{87.0} & 88.0 & \textbf{75.0} & 78.0 \\ \hline

SVM rbf & t1-t3 & 78.9 & \textbf{85.2} & 70.5 & \textbf{81.7} \\
    & t1-t5 & \textbf{82.6} & 82.6 & \textbf{95.8} & 74.6 \\ \hline

\end{tabular}%
}
\end{table}

\begin{figure} [htb]
\centering
\includegraphics[width=\textwidth]{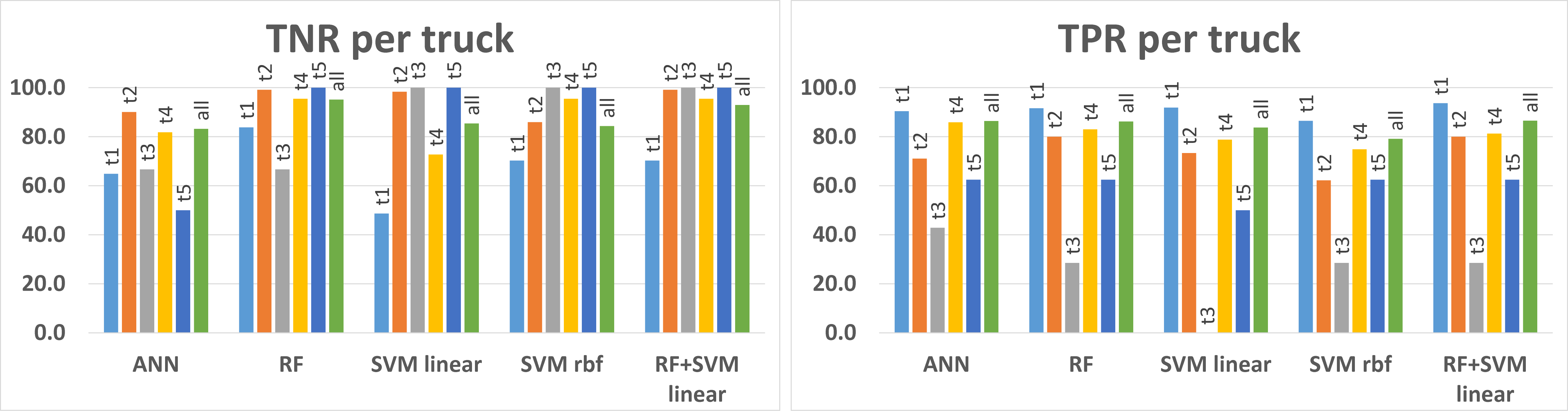}
\caption{\textit{TNR} and \textit{TPR} values of the classifiers per truck when the classifiers are trained with data from trucks t1-t5. Better seen in colour.}
\label{fig:TNR_TPR_per_truck_and_fusion}
\end{figure}

\begin{table}[]
\centering
\caption{\textit{TNR} and \textit{TPR} values of the classifiers per truck when the classifiers are trained with data from trucks t1-t5. The best cases per column are marked in bold.}
\label{tab:TNR_TPR_per_truck_and_fusion}
\resizebox{\textwidth}{!}{%
\begin{tabular}{|c|cccccc|cccccc|}

\cline{2-13}

\multicolumn{1}{c|}{} & \multicolumn{6}{c|}{\textbf{TNR}} & \multicolumn{6}{c|}{\textbf{TPR}} \\  \hline

\textbf{classifier} & t1 & t2 & t3 & t4 & t5 & all & t1 & t2 & t3 & t4 & t5 & all \\ \hline

ANN & 64.9 & 90.1 & 66.7 & 81.8 & 50.0 & 83.2 &  90.4 & 71.1 & \textbf{42.9} & \textbf{85.9} & \textbf{62.5} & 86.4  \\ \hline

RF & \textbf{83.8} & \textbf{99.2} & 66.7 & \textbf{95.5} & \textbf{100.0} & \textbf{95.1} &  91.6 & \textbf{80.0} & 28.6 & 83.0 & \textbf{62.5} & 86.2  \\ \hline

SVM linear & 48.6 & 98.3 & \textbf{100.0} & 72.7 & \textbf{100.0} & 85.4 &  91.9 & 73.3 & 0.0 & 78.8 & 50.0 & 83.7  \\ \hline

SVM rbf & 70.3 & 86.0 & \textbf{100.0} & \textbf{95.5} & \textbf{100.0} & 84.3 &  86.5 & 62.2 & 28.6 & 74.9 & \textbf{62.5} & 79.1 \\ \hline \hline

RF + SVM linear & 70.3 & \textbf{99.2} & \textbf{100.0} & \textbf{95.5} & \textbf{100.0} & 93.0 &  \textbf{93.7} & \textbf{80.0} & 28.6 & 81.3 & \textbf{62.5} & \textbf{86.5} \\ \hline 
 
\end{tabular}%
}
\end{table}

\subsection{Per-truck analysis and classifier fusion}

We finally show the performance per truck, considering the configuration of the classifiers of the last sub-section (trained with both \textit{mirror} and \textit{CMS} trucks).
Results are given in Figure~\ref{fig:TNR_TPR_per_truck_and_fusion} and Table~\ref{tab:TNR_TPR_per_truck_and_fusion}.

As can be seen, the performance varies between trucks.
\textit{TPR} shows a consistent relative tendency across all classifiers, with t1 performing the best, followed by t4, t2, t3, and t5. 
Notice that this performance trend correlates with the size of the overtake class in the training/test sets (Table~\ref{tab:db-training-test-files}). 
Despite the classifiers being trained with balanced classes, t3 and t5 just have 3-4 training files per truck. 
This smaller sample size for these trucks may mean that the classifiers are not sufficiently trained on them.
Although driver or truck model information is not available, each truck is likely driven by a different person. 
Even if overtake patterns were observed to be consistent (Figures~\ref{fig:boxplots_mirror} and \ref{fig:boxplots_CMS}),
small individual differences per driver or truck type may occur, which could impact the \textit{TPR} given the smaller training sample size for t3 and t5.

\textit{TNR}, on the other hand, is not consistent across classifiers and trucks. For example, t3 has a poor \textit{TNR} with RF and a good \textit{TNR} with SVM linear, while t4 shows the opposite pattern. 
This inconsistency may arise from the heterogeneous nature of the no-overtake class. Our dataset includes a wide range of driving situations in such class, such as turns, avoiding an exit-only lane, giving way to a vehicle, surpassing stationary vehicles, and others. However, we have not systematically captured these additional possibilities, as the dataset comes from real-world trucks in regular operation, where such events can occur randomly.
%
%
An obvious solution would be to expand the dataset, thereby improving the representativeness of the various non-overtake manoeuvres.

%
%

Given such different \textit{TNR} performance per classifier and per truck, we tried classifier fusion by averaging their output scores. 
We tested all possible combinations of 2, 3 and 4 classifiers. The most consistent was observed to be Random Forest combined with SVM linear, whose results are provided in Figure~\ref{fig:TNR_TPR_per_truck_and_fusion} and Table~\ref{tab:TNR_TPR_per_truck_and_fusion} too. 
As shown in the plot, the fusion boosts \textit{TNR} of most trucks, with the exception of t1, which experiences a decline. 
\textit{TPR}, on the other hand, does not improve with fusion (except for t1, where \textit{TNR} worsened).
Overall, while the fusion does not impact \textit{TPR}, it helps to balance and improve \textit{TNR}, and as observed in the table, the fusion yields the best performance across most trucks.

\section{Conclusions}
\label{sect:conclusions}

Safe truck overtakes are crucial to avoid collisions, reduce congestion, and guarantee smooth traffic flow.
Their reliable prediction can enable advanced Driver Assistance Systems (ADAS) to provide timely, informed driving decisions, ultimately enhancing road safety and efficiency.
%
%
This study investigates the capabilities of machine learning classifiers trained on CAN (Controller Area Network) signals to detect overtakes in trucks.
In particular, we have applied traditional, widely used classifiers in related tasks on vehicle manoeuvre detection \citep{xing2019driver,kim2017prediction,das2020detecting}, such as Support Vector Machines (SVM), Random Forest (RF), and Artificial Neural Networks (ANN) 
For this study, we gathered CAN data from actual operating trucks provided by the Volvo Group participating in this research.
To the best of our knowledge, we are among the first to address overtake detection in trucks, particularly using real CAN bus data.

To collect data, we designed a precondition trigger representative of an overtake, so the data logger only records when a certain combination of CAN signals occurs. This prevented storage problems and allowed us to obtain segments where a possible overtake occurs.
Data from a dashcam was also captured, which allowed offline manual labelling of the obtained segments.
With this procedure, we obtained 865 overtake segments and 382 no-overtake segments from 5 different trucks. 
The precondition trigger was designed to filter out city traffic events (below 50 km/h) that are not really overtakes, or at least not as dangerous as an overtake at high speeds, so the resulting files mostly correspond to highways or non-urban roads.

Throughout different experiments, we then demonstrate the suitability of CAN bus data to detect overtakes in trucks.
We have examined a variety of configurations (Figure~\ref{fig:system}) that involve, for example, cropping the files around the precondition trigger with different time lengths, processing CAN signals with a sliding window of different lengths, or extracting various metrics within the window, such as the mean or the standard deviation of the signals.
This is because the value of such parameters is subject to discussion in the related literature \citep{xing2019driver,khairdoost2020real}.
Relevant findings include, for example, that taking a very large crop before or after the trigger is detrimental to detecting overtakes. 
Our optimal segment crop comprises 5 seconds before the trigger and just 1 second afterwards.
Thus, the most helpful information for detecting the overtake class becomes more evident a few seconds before the trigger and disappears shortly after.
This somehow limits the capability of the system to detect overtakes reliably to just a few seconds before it starts.

Other findings concern the optimal sliding window size or the metrics to be gathered within the window, which is observed to be different per classifier.
We have also analysed the pattern of CAN signals to understand their dynamics. 
The behaviour of drivers during overtakes appears to be consistent across different trucks, with many signals remaining consistent. 
Two of the trucks have a mirrorless Camera Monitoring System (CMS), while the other three have a standard physical mirror. Although it was not our primary aim in this paper to compare them, we have not observed any remarkable difference in the pattern of overtake signals between the two truck types, suggesting that the absence of a physical mirror does not substantially affect the way that drivers carry out an overtake.
On the other hand, differences are observed in no-overtake manoeuvres. 
Data from some files is observed to be collected under heavier traffic conditions than others, which results in a different pattern of CAN signals.
However, we have not systematically controlled such scenarios. 
Our data was captured with actual operating trucks normally driven, so the driving moments where the trigger activates can occur randomly.
The effect is that while the Positive Rate of the true class (\textit{TPR}) remains consistent, the True Negative Rate (\textit{TNR}) is highly dependent on the presence of different non-overtake patterns in the training data.

In the last part of our study, we analysed performance at the individual truck level. For the overtake class, we observed a high dependency on the amount of training data available per truck, regardless of the classifier. This suggests that subtle differences per driver or truck type may also introduce inconsistencies in the training set if they are not adequately represented.
On the contrary, the no-overtake class exhibited less consistency. We did not observe a clear dependency on the size of the training set or the classifier type. We believe that the non-systematic nature of the data acquisition process mentioned earlier may produce such a lack of consistency. An obvious approach would be to increase the database, considering sufficient representativity of possible non-overtake manoeuvres.   
Another way that we have found to counteract such an effect is classifier fusion.
Through simple score fusion of two classifiers (RF and SVM linear), the \textit{TNR} of most trucks is boosted to more than 95\%, without the \textit{TPR} being significantly impacted.
Overall, the fusion provides a \textit{TNR} of 93\% for all trucks, and a \textit{TPR} of 86.5\%.

\subsection{Limitations and Future Work}

In the context of ADAS, both false positives (related to TNR$<$100\%) and false negatives (TPR$<$100\%) carry practical safety implications. 
%
%
%
The potential impact of these errors depends on how the detection module is integrated into the ADAS pipeline. 
For real-time systems, false positives can be particularly problematic, since erroneous alerts may distract or confuse the driver, especially if they happen repeatedly. If the system is designed to adapt vehicle behaviour (e.g., adjusting speed, spacing, or lane position), it could cause unwarranted actions, potentially increasing risk instead of reducing it. Ultimately, false positives can erode driver trust, leading to disengagement or disregard of future valid alerts.
In contrast, false negatives (undetected overtakes) in real-time reduce the system's ability to provide timely support or warnings. It may be acceptable in non-critical situations or when the driver is in full control already. 
However, while less disruptive than false positives, frequent false negatives may affect the usefulness of the system in critical situations or degrade the coverage of the ADAS functionality.
On the other hand, in offline analysis, such as post-trip analysis, driver profiling, or fleet safety auditing, classification errors have less immediate impact.
False positives typically lead to the mislabeling of routine driving as overtaking, slightly skewing statistics but without real-time consequences. 
False negatives reduce completeness but rarely compromise safety. 
Nonetheless, systematic errors or biases can still distort behavioural summaries or affect managerial decision-making. 
Therefore, maintaining high reliability remains important even in non-real-time pipelines.

In terms of real-time feasibility, although not directly evaluated in this study, the methods used are based on lightweight models (ANN, RF, SVM) and low-dimensional CAN signal statistics, which are well-suited for embedded implementation. 
In future work, we plan to port the trained classifiers to onboard platforms and evaluate inference times and resource consumption. Given the simplicity of the models and the preprocessing pipeline, we expect that real-time performance can be achieved with minimal computational overhead.
In addition, all three classifiers can be trained efficiently due to moderate input dimensionality. RF trains within seconds. SVM, though more computationally demanding, keeps training times between seconds and minutes on a standard workstation for linear kernels. The ANN used has a simple architecture (one hidden layer), and training converges in less than one minute.
Beyond technical performance, our system has managerial and industrial relevance.
Since it relies solely on CAN bus signals, it can be deployed without requiring additional sensors, making it a cost-effective, scalable, and privacy-preserving solution suitable for fleet-wide deployment.
In addition to potential real-time safety applications, the system can support long-term driver behaviour monitoring and profiling, safety auditing, insurance risk assessment, and driver training programs by identifying risky patterns or habits.

Several factors may contribute to classification errors, including variability in traffic context, differences in driving styles, sensor noise, and the limited number of labelled overtaking events, especially in some trucks. These factors may result in signal patterns that deviate from typical examples seen during training.
To mitigate these risks and improve the reliability of the system in operational settings, future implementations may incorporate more sophisticated temporal smoothing techniques (e.g. decision aggregation over consecutive windows of shorter or longer duration, that also go earlier in time), combined with variable confidence-based thresholds, so the evidence is accumulated over time as the overtake event approaches, rather than relying on the output of a single crop. 
Hybrid sensor fusion could be another possibility, combining CAN-based predictions with camera or radar inputs, although it goes beyond our purpose of employing only CAN signals. 

The above experiments have been carried out with the majority of the parameters of the classifiers set to default. 
Some observations may also change with another parametrisation, e.g. via classifier optimisation \citep{10.5555/3020751.3020778} or parameter selection \citep{8735037}, that could be more suitable for our type of data. 
In addition, a way to increase the detection window of an overtake earlier in time could be to extend the database, which would also enable the use of data-hungry popular deep learning models such as Long Short-Term Memory (LSTM) networks \citep{zhang2020hybrid,khairdoost2020real} or Transformers \citep{guo2022lane}.
While we focused on traditional machine learning methods due to their robustness in low-data scenarios, we acknowledge that such advanced deep learning architectures may offer improved performance for time-series data like CAN signals.
%
%
Contrary to the algorithms of this paper, such models exploit temporal dependencies in sequential data, capturing more intricate patterns in the signals, long-range dependencies and hierarchical temporal patterns, potentially providing a broader window and better performance. 
Recent advances in large language models (LLMs) open new possibilities for multivariate time-series analysis, including vehicle sensor data, due to the LLM's potential superior accuracy compared to traditional machine learning and even deep-based models such as CNNs and LSTMs \citep{10843617}. 
Although LLMs are primarily designed for textual data, their underlying architecture, the Transformer, has shown strong performance in time-series applications through variants like Informer \citep{haoyietal-informer-2021}, Autoformer \citep{wu2021autoformer}, and more recently, TST, PatchTST \citep{Yuqietal-2023-PatchTST} and LLM4TS \citep{10.1145/3719207}.

These models, however, typically require significantly larger volumes of labelled data to train effectively, which was a limiting factor in our current setup.
In this future work direction, we are working on capturing large amounts of unlabeled data from the employed trucks by recording driving sessions continuously. This would allow to increase our dataset, for example, via pseudo-labelling \citep{li2019naive}, selecting samples with high prediction probability as given by the classifiers trained with labelled data, thus enabling the use of the mentioned deep models.
A larger dataset would also increase the diversity of signal patterns in non-overtake manoeuvres, identified as one of the issues emerging from the realistic and unconstrained nature of our data acquisition, where class diversity cannot be guaranteed a priori.

%
A unique part in future phases of the BIG FUN project will be the extension of the dataset with multimodal data, including video streams from digital cameras placed inside and around the truck, complementing traditional data such as CAN signals and Driver Interaction Input data. This will allow us to build a more holistic representation of truck usage and driver behaviour across different contexts. These multimodal data sources will be combined with AI-based analysis to automatically identify moments of potential significance in truck journeys, not only overtakes. Coupled with user experience (UX) studies and expert evaluation, these AI-discovered events are expected to generate actionable insights that contribute to safer, more intelligent vehicle systems and improved commercial mobility.

\section*{CRediT authorship contribution statement}

\textbf{Fernando Alonso-Fernandez:} 
Conceptualization,
Writing – original draft,
Writing – review \& editing,
Methodology,
Software,
Validation,
Supervision,
Funding acquisition.
\textbf{Talha Hanif Butt:} 
Writing – original draft,
Writing – review \& editing,
Data curation,
Software.
\textbf{Prayag Tiwari:} 
Conceptualization,
Writing – review \& editing,
Methodology,
Validation,
Supervision.

\section*{Declaration of Competing Interest}
The authors declare that they have no known competing financial interests or personal relationships that could have appeared to influence the work reported in this paper.

\section*{Acknowledgements}
The authors thank the BIG FUN (Big Data-Powered End User Function Development) project of the Swedish Innovation Agency (VINNOVA) for funding this research (project 2021-05045).
The authors would like to acknowledge Volvo Group for their collaboration and support in providing access to in-service truck data, which was instrumental in conducting this research.


\section*{Declaration of generative AI and AI-assisted technologies in the writing process}

During the preparation of this work, the authors used ChatGPT in order to proofread the text and improve its readability. After using this tool/service, the authors reviewed and edited the content as needed and take full responsibility for the content of the publication.

\bibliography{refs}

\end{document}